\PassOptionsToPackage{unicode}{hyperref}
\PassOptionsToPackage{hyphens}{url}
\documentclass[
]{article}
\usepackage{amsmath,amssymb}
\usepackage{lmodern}
\usepackage[utf8]{inputenc}
\usepackage{iftex}
\usepackage{relsize}

\ifPDFTeX
  \usepackage[T1]{fontenc}
  \usepackage[utf8]{inputenc}
  \usepackage{textcomp} 
\else 
  \usepackage{unicode-math}
  \defaultfontfeatures{Scale=MatchLowercase}
  \defaultfontfeatures[\rmfamily]{Ligatures=TeX,Scale=1}
\fi
\IfFileExists{upquote.sty}{\usepackage{upquote}}{}
\IfFileExists{microtype.sty}{
  \usepackage[]{microtype}
  \UseMicrotypeSet[protrusion]{basicmath} 
}{}
\makeatletter
\@ifundefined{KOMAClassName}{
  \IfFileExists{parskip.sty}{%
    \usepackage{parskip}
  }{
    \setlength{\parindent}{0pt}
    \setlength{\parskip}{6pt plus 2pt minus 1pt}}
}{
  \KOMAoptions{parskip=half}}
\makeatother
\usepackage{xcolor}
\usepackage{graphicx}
\usepackage{parskip}
\usepackage{longtable,booktabs,array}
\usepackage{calc} 
\usepackage{etoolbox}
\makeatletter
\patchcmd\longtable{\par}{\if@noskipsec\mbox{}\fi\par}{}{}
\makeatother
\IfFileExists{footnotehyper.sty}{\usepackage{footnotehyper}}{\usepackage{footnote}}
\makesavenoteenv{longtable}
\makeatletter
\def\maxwidth{\ifdim\Gin@nat@width>\linewidth\linewidth\else\Gin@nat@width\fi}
\def\maxheight{\ifdim\Gin@nat@height>\textheight\textheight\else\Gin@nat@height\fi}
\makeatother
\setkeys{Gin}{width=\maxwidth,height=\maxheight,keepaspectratio}
\makeatletter
\def\fps@figure{htbp}
\makeatother
\ifLuaTeX
  \usepackage{selnolig}  
\fi
\IfFileExists{bookmark.sty}{\usepackage{bookmark}}

\IfFileExists{xurl.sty}{\usepackage{xurl}}{} 
\hypersetup{
  hidelinks,
  pdfcreator={LaTeX via pandoc}}

\usepackage{xcolor}
\usepackage{hyperref}
\hypersetup{
  colorlinks,
  citecolor=blue,
  linkcolor=black,
  urlcolor=blue}

\author{}
\date{}

\usepackage{blindtext}
\usepackage{geometry}
\usepackage{natbib}
\setcitestyle{authoryear,open={},close={}}

\usepackage{moresize}

\begin{document}
\begin{center}

\large \textbf{A Theory of Intelligences}

\normalsize Michael E. Hochberg

ISEM, Université de Montpellier, CNRS, IRD, EPHE, 34095 Montpellier,
France

Santa Fe Institute, Santa Fe, NM 87501, USA

\href{mailto:Michael.Hochberg@UMontpellier.fr}{\nolinkurl{Michael.Hochberg@UMontpellier.fr}}

\today

\hfill \break

\end{center}

\hypertarget{Abstract}{%
\section{Abstract}\label{Abstract}}

Intelligence is a human construct to represent the ability to achieve
goals. Given this wide berth, intelligence has been defined countless
times, studied in a variety of ways and represented using numerous
measures. Understanding intelligence ultimately requires theory and
quantification, both of which have proved elusive. I develop a framework -- the Theory of Intelligences (TIS) -- that applies across all systems from physics, to biology, humans and AI. TIS likens intelligence to a calculus, differentiating, correlating and integrating information. Intelligence operates at many levels and scales and TIS distils these into a parsimonious macroscopic framework centered on solving, planning and their optimization to accomplish goals. Notably, intelligence can be expressed in informational units or in units relative to goal difficulty, the latter defined as complexity relative to system (individual or benchmarked) ability. I present general equations for intelligence and its components, and a simple expression for the evolution of intelligence traits. The measures developed here could serve to gauge different facets of intelligence for any step-wise transformation of information. I argue that proxies such as environment, technology, society and collectives are essential to a general theory of intelligence and to possible evolutionary transitions in intelligence, particularly in humans. I conclude with testable predictions of TIS and offer several speculations. 
\\

\textbf{Keywords}: Benchmark, Calculus, Complexity, Difficulty, 
Embodiment, Efficiency, Evolution, Goals, Information, Intelligence niche, Knowledge, Metacognition, 
Optimization, Planning, Solving, Surprisal, System, Uncertainty

\newpage
\hypertarget{introduction}{%
\section{INTRODUCTION}\label{introduction}}

The famous quote ``information is the resolution of uncertainty'' is
often attributed to Claude Shannon. What Shannon's seminal 1948 article
``A mathematical theory of communication'' and the above quote indicate
is that information is a potential: unresolved order
that can be perceived, filtered, and deciphered. Shannon had previously
equated information with the term ``intelligence'' in his 1939
correspondence with Vannevar Bush (\citep{rogersClaudeShannonCryptography1994a}), though his observation is but one essential part of intelligence, the other being
goal attainment (\citep{leggCollectionDefinitionsIntelligence2007}). By associating these concepts,
one arrives at a simple, general description of intelligence: the
resolution of uncertainty producing a result or goal. My objective is
to unpack this statement towards a general, inclusive definition of
intelligence and use this to propose a series of increasingly complete frameworks and a theory of intelligence.

Intelligent entities have one or more models of the world (e.g., \citep{albusOutlineTheoryIntelligence1991a}; \citep{hawkinsThousandBrainsNew2021}; \citep{lecunPathAutonomousMachine2022}). Simply
possessing a model however does not necessarily invoke what most humans would regard as intelligence. Thus,
systems from atoms to molecules to gasses, liquids and solids are all
governed by physical laws -- models of a sort -- but these laws are reactive: except under special circumstances, gasses, liquids and solids cannot generate structure
spontaneously and they do not have the goal directedness usually associated
with (human) intelligence. Nevertheless, the laws of physics and chemistry can
be driven to a local reduction in entropy, that is, decreased
uncertainty (\citep{brillouinNegentropyPrincipleInformation1953}) and therefore a most bare-bones
instantiation of intelligence -- the reorganization or generation of
information (\citep{landauerInformationPhysical1991}). However, what is notably missing from
physical intelligence that biological entities do have is alternatives
to pure reaction. That
is, natural physical systems do not make goals, nor can they actively
choose among alternative paths or check and rectify errors towards goals. Reducing decision and prediction errors at different organizational
levels is part of what differentiates thinking from non-thinking systems
such as computers and AI (\citep{tononiConsciousnessHereThere2015}; \citep{hohwySelfEvidencingBrainSelfEvidencing2016}).
Computers and artificial intelligence -- even if capable of impressive
feats from a human perspective -- have models (e.g., software,
algorithms) that are still far simpler than biological systems and
humans in particular (\citep{roitblatAlgorithmsAreNot2020a}). Among the main features added in
the huge and fuzzy steps from machine to human intelligence are
environmental sensitivity and active inference (\citep{kortelingHumanArtificialIntelligence2021}).

These and other complexities present a major challenge to developing a
theory of intelligence (\citep{duncanComplexityCompositionalityFluid2017b}; \citep{barbeyNetworkNeuroscienceTheory2018}; \citep{stemlerRaschMeasurementItem2021}) in both biological (\citep{epsteinEmptyBrain2016}) and artificial
(\citep{carabantesBlackboxArtificialIntelligence2020}; \citep{bodriaBenchmarkingSurveyExplanation2021}) systems. Intelligence theory has largely
focused on humans, identifying different milestones from simple reactions
through to the more elaborate, multi-level processes involved in
thought. Raymond Cattell distinguished acquired and active components of
intelligence in humans, defining crystallized intelligence as the
ability to accumulate and recall knowledge (much like a computer), and
fluid intelligence as abilities to learn new skills and to apply
knowledge to new situations (a thinking entity) (\citep{cattellTheoryFluidCrystallized1963}).
Although an oversimplification of the many factors and interactions
forming intelligence (\citep{mcgrewCattellHornCarrollCognitiveachievementRelations2010}; \citep{flanaganCattellHornCarrollTheoryCognitive2014}), this basic dichotomy is useful in differentiating the functional
significance of storage/recall to familiarity versus active decision making when faced with novel situations (\citep{parrActiveInferenceFree2022a}). The more recent \textit{network neuroscience theory} takes a mesoscopic
approach in linking network structure with memory and reasoning
components of intelligence (\citep{barbeyNetworkNeuroscienceTheory2018}), but is impractical for
dissecting exhaustive pathways and more efficient short-cuts.
Other theories of intelligence have attempted to integrate
either processes, emergent phenomena or both (e.g., \citep{sternbergTriarchicTheoryHuman1984}; \citep{albusOutlineTheoryIntelligence1991a}; \citep{goertzelStructureIntelligenceNew1993}; \citep{wangAbstractIntelligenceUnifying2009}; \citep{georgiouPASSTheoryIntelligence2020}). Arguably, the main shortcoming of current theory is
the view that the reference for intelligence is thinking entities, thus
ignoring structural features linking physics, different milestones in
biology and artificial systems, and, with notable exceptions (\citep{flinnEcologicalDominanceSocial2005}; \citep{rothEvolutionBrainIntelligence2005}; \citep{sterelnySocialIntelligenceHuman2007}; \citep{readerEvolutionPrimateGeneral2011}; \citep{burkartEvolutionGeneralIntelligence2017}; \citep{shreeshaCellularCompetencyDevelopment2023}), ignoring the roles of transmission and evolution in intelligence.

I develop a theory that intelligence is a fundamental property of \textit{all} systems and is exhibited in a small number of distinct, distinguishing forms.
Previous research has explained intelligence in levels or hierarchies
(e.g., \citep{spearmanGeneralIntelligenceObjectively1904}; \citep{albusOutlineTheoryIntelligence1991a}; \citep{carrollHumanCognitiveAbilities1993}; \citep{conradCrossscaleInformationProcessing1996}; \citep{adamsMappingLandscapeHuman2012}; \citep{dearyStabilityIntelligenceChildhood2014}; \citep{cholletMeasureIntelligence2019}; \citep{lecunPathAutonomousMachine2022}; \citep{burgoyneAttentionControlProcess2022}; \citep{fristonDesigningEcosystemsIntelligence2024}) and I employ this idea to account for the variety of
intelligences, including natural physical systems, biological ones and
humans in particular, and artificial and designed systems. I define a ``system'' (also referred to below as an ``agent'')
as an object that uses external energy sources to lower
entropy over one or more of a defined function, an object, space or time. Generic examples include
biological development, generating information, and maintaining homeostasis. 

The Theory of
Intelligences (TIS) is developed in three parts,
starting with temporal goal resolution, then recognising micro-, meso- and
macroscopic abilities, and finally accounting for
system-life-long changes in intelligence and transmission through time,
including the evolution of intelligence traits. The key advances of TIS
are (1) the partitioning of intelligence into local uncertainty reduction
(``solving'') and global optimization (``planning''); (2) distinguishing
challenges in the forms of goal difficulty and surprisal; (3)
recognizing not only the core system, but extra-object spaces, including past sources, present proxies (i.e., any support that is not part of a system at its inception), environments, present and near-future transmission, and distant evolution.

As a start towards a formalization of TIS, I present mathematical
expressions based on the quantifiable system features of
solving and planning, difficulty, and optimal (efficient, accurate and complete) goal resolution. Goals may be imposed by necessity, such as survival imperatives, and/or be opportunistic or actively defined by the system, such as preferences, learning new skills, or goal definition itself. Solving and planning have been extensively discussed in the intelligence literature, and the advance of TIS is to mathematically formalize their contributions to intelligence and to represent how, together with optimization, they constitute a parsimonious, general theory of intelligence. The proposed partitioning of solving and planning is particularly novel
since it predicts that paths to a goal not only function to achieve
goals, but also may constitute experimentations leading to higher probabilities for
future attainable goals and increased breadth to enter new goal spaces, moreover possibly serving as a generator of variations upon which future selection will act.
These experimentations moreover are hypothesized to explain capacities and
endeavors that do not directly affect Darwinian fitness, such as
leisure, games and art. I do not discuss in any detail the many theories of intelligence nor the quantification of
intelligence, the latter for which the recent overview by
Hernández-Orallo (\citep{hernandez-oralloMeasureAllMinds2017}) sets the stage for AI, but
also yields insights into animal intelligence and humans. Neither do I
discuss the many important, complex phenomena in thinking systems such as cognition, goal directedness and
agency (\citep{babcockGoalDirectednessField2023}).

\hypertarget{THE IDEA}{%
\section{THE IDEA}\label{THE IDEA}}

My argument is that goals -- be they results or the means to obtain results -- are informational constructs, and
because such assemblies and their component sets are potentially
complex (modular, multidimensional, multi-scale), two fundamental
capacities potentially contribute to optimally (efficiently, accurately and/or completely) attain resolutions. These are: (1) the ability to resolve uncertainty (`solving') and (2) the ability to partition a complex goal into a sequence of subgoals (`planning') (\hyperlink{fig1}{Figure 1}). 

Briefly, both solving and planning harness priors, knowledge
and skills to accomplish goals, but the
former focuses on the myopic resolution of goal elements, whereas the
latter is the broader assessment of alternative steps to
attain goals. Planning is a manifestation of the ``adjacent possible'' (\citep{kauffmanInvestigations2000}), whereby the cost of ever-future horizons is (hyper)exponentially increasing uncertainty. Not surprisingly, planning is expected to require more working memory (and for hierarchical planning, causal understanding) than
does solving (\citep{albusOutlineTheoryIntelligence1991a}; \citep{duncanComplexityCompositionalityFluid2017b}). Planning could be influenced by a system's perceived
ability to solve individual subgoals (including how particular choices affect future subgoals), though this does not mean that
planning is necessarily harder than solving. Moreover, excellent solving
ability alone could be sufficient to attain goals (i.e., solving is both
necessary and sufficient for some goal/system types), whereas the capacity to
identify the most promising path alone may or may not be necessary and is \emph{not sufficient} to
resolve a goal (\hyperlink{fig1}{Figure 1A}). The value added of the latter capacity is increased
accuracy, precision (hereafter precision will be lumped
into the related term, accuracy), efficiency and completeness, and these become more difficult
to achieve as goals become increasingly complex and therefore difficult to represent or understand (\hyperlink{fig1}{Figure 1B}). 

Based on the above observations, a sequence in the emergence and
evolution of intelligence \emph{must} begin with resolving uncertainty
(solving) and thereafter possibly expand into the \emph{relevance} of
alternative sequential informational sequences (planning). This is the basis for evolutionary reasoning to explain
differences in intelligence capacities between individual systems
(trait variation, environment) and across system types (phylogeny), with physical
systems at the base and a hypothetical hierarchy in complexity
as one goes from artificial (and subdivisions) to biological (and
subdivisions) to human. I stress that organizing systems in intelligence
classes does not signify differences in performance (quality, value or superiority). Rather, it relates to how
these different systems adapt to their particular spectrum of
environmental conditions and goals. Differences in the
amplitude and spectrum of intelligence traits are therefore hypothesized to
reflect differences in the \textit{intelligence niche}, that is
environments, capacities and goals relevant to a system (e.g.,
\citep{godfrey-smithEnvironmentalComplexityEvolution2002}; \citep{pinkerCognitiveNicheCoevolution2010}; \citep{burkartEvolutionGeneralIntelligence2017}).

Key to my idea is that fitness, information, complexity, entropy,
uncertainty and intelligence are interrelated. There is abundant
precedent for such associations, going at least back to Shannon (\citep{shannonMathematicalTheoryCommunication1948}), based on
comparisons between subsets of the six phenomena (e.g., \citep{gottfredsonWhyMattersComplexity1997}; \citep{donaldson-matasciFitnessValueInformation2010}; \citep{frankNaturalSelectionHow2012}; \citep{adamiElementsIntelligence2023}). One of my
objectives is to begin to explore these interrelations, recognizing that
an ecological perspective could yield important and general insights.
Thus, intelligence crosses scales, for example, from phenomena occurring
within systems to interactions between systems and their environment, to
interactions in populations of systems. Scales and the orthogonal
concept of levels introduce notions of complexity, that is,
heterogeneities in one or more influential factors or structures (\citep{mcsheaPerspectiveMetazoanComplexity1996}; \citep{godfrey-smithEnvironmentalComplexityEvolution2002}). I hypothesize that the hierarchical nature of
complexity at different scales is a manifestation of innovation and more singular \emph{transitions in intelligence}, that is, from a baseline of the resolution of
local uncertainty (all systems), to sequential relations among local
uncertainties (most biological systems to
humans), to the ability to integrate two or more local uncertainties so
as to more accurately, completely and efficiently achieve goals (higher
biological cognitive systems to humans) and
finally general intelligence in humans. Although largely unexplored (but
see \citep{stanleyCompetitiveCoevolutionEvolutionary2004}; \citep{baluskaHavingNoHead2016}; \citep{frankIntelligencePlanetaryScale2022}; \citep{fieldsCompetencyNavigatingArbitrary2022a}), I suggest that the transitions from baseline uncertainty
resolution to general intelligence reflect increased solving ability and for more complex goals, hierarchical planning, including its integration with basic-level solving.

That complexity and intelligence can be intertwined has important
implications for explaining structure and function across
systems. Insofar as systems evolve, so too do the intelligence traits
employed to penetrate existing complexity (e.g., define and realize
goals; \citep{demartinoGoalsUsefulnessAbstraction2023}) and contribute to changes in complexity (e.g., generate
novel information, structures, transitions in scales and levels), which, in turn, require greater and different forms of intelligence in order
to resolve, and so on, possibly leading to an auto-catalytic process. Goal relevancy
includes those actions affecting fitness (i.e., growth, survival and
reproduction when faced with challenging environments, including limited
resources, competitors and predators), well-being (e.g., sports, leisure,
art in humans), or actions with no apparent objective at all. The goal can
be within the gamut of previous experiences or an extension thereof, or
be novel but realizable at least in part, despite its difficulty. To the extent
that the universe of relevant, feasible goals is diverse in
complexities, one expectation is that intelligence traits will not only
evolve to improve fitness in existing niches, but also extend into new
intelligence niches, which may or may not be more complex than existing
ones (\hyperlink{fig2}{Figure 2}) (for ecological niche concepts see \citep{chaseEcologicalNichesLinking2003}).

In sum, we need parsimonious theory that recognizes the close interconnections
between uncertainty, information, complexity and intelligence, and is built on
ecological and evolutionary principles.

\hypertarget{A BASIC FRAMEWORK}{%
\section{A BASIC FRAMEWORK}
\label{A BASIC FRAMEWORK}}

The above discussion equates intelligence with a system (or a collection of
systems) reacting to or engaging in a challenge or an opportunity, and to
do so, perceiving, interpreting, manipulating and assembling information
in its complexity, and rendering the information construct in a
different, possibly more complex form. An example
of a complex finality is an architecturally novel, high-level
functioning business center. An example of a simple goal outcome is to win in
chess. Both examples may have highly challenging paths, and may or may not differ
in the information actually contained in the finality.

Perceiving, interpreting, manipulating and assembling employ logical
piece-wise associations (\citep{albusOutlineTheoryIntelligence1991a}; \citep{lecunDeepLearning2015}; \citep{lecunPathAutonomousMachine2022}) and as such at a
more abstract, computational level, they differentiate, correlate and
integrate raw data and existing goal-useful information (\hyperlink{fig3}{Figure 3}). This involves amassing and
deconstructing complex and possibly disjoint ensembles,
identifying their interrelationships, and reconstructing the ensembles
and existing knowledge as goal-related information, so as to infer or
deduce broader implications leading to a resolution (\citep{adamiElementsIntelligence2023}). Thus,
intelligence is an operator or a \emph{calculus of information}.

The above coarse-grain processes emerge from more microscopic, fine-grained
 levels with what are often viewed as traits associated with humans such as
reasoning, abstraction, generalization and imagination. Unfortunately, there is no
single objective way to mathematically represent these and other
microscopic features of intelligence and their interrelationships (e.g., \citep{albusOutlineTheoryIntelligence1991a}; \citep{mccarthyProposalDartmouthSummer2006}; \citep{gershmanComputationalRationalityConverging2015}; \citep{lecunDeepLearning2015}; \citep{scholkopfCausalRepresentationLearning2021}; \citep{adamiElementsIntelligence2023}). Thus, to be manageable and useful, a general
theory of intelligence needs to be based on higher-level
macroscopic observables. To accomplish this, I begin by developing a series of
increasingly complete conceptual models.

\newpage
Consider the following keyword sequence for intelligence:
\large
\begin{center}
\textbf{A: Use Prior Information Current Data Towards Goal Result}
\end{center}
\normalsize
Despite overlap between these descriptors, \textbf{A} makes plain the
temporal nature of intelligence. The process requires integrating previously
acquired information in the forms of priors and experience,
including knowledge and skills, and the use of current data and information in a world
model towards a future goal, producing a result. Key is the manipulation
and application of raw data and more constructed information towards an
objective.

To integrate these elements into a general process we first focus on the
core processing module:

\begin{center}
\textbf{B: Goal} \(\rightarrow^{y}\) \textbf{Path} \(\rightarrow^{z}\)
\textbf{Result}
\end{center}

System \textbf{B} is general. If \textbf{B} is a mere reporter such as a
computer, then it simply uses input to find output from an existing
list, or Goal → Result. If a system with more elaborate capacities, then \(\rightarrow^{y}\) may be characterized by signal processing, simulation,
interpretation and eventual reformulation of the problem, followed by
engagement in the decided method of resolution to \(\rightarrow^{z}\), including
prediction and checking for errors (\citep{lecunDeepLearning2015}). This
more elaborate system then decides whether to go back to \(\rightarrow^{y}\) and possibly
use what previously appeared to be useless data (or faulty
computational methods), or continue on and render a result.
Understanding intelligence in \textbf{B} thus requires we have
information about its complex inner workings (\citep{croninImitationGameComputational2006}).

We can further generalize \textbf{B} to how information is accessed,
stored, processed and used towards a goal:
\begin{figure}	
\begin{center}
    \includegraphics[width=0.5\textwidth]{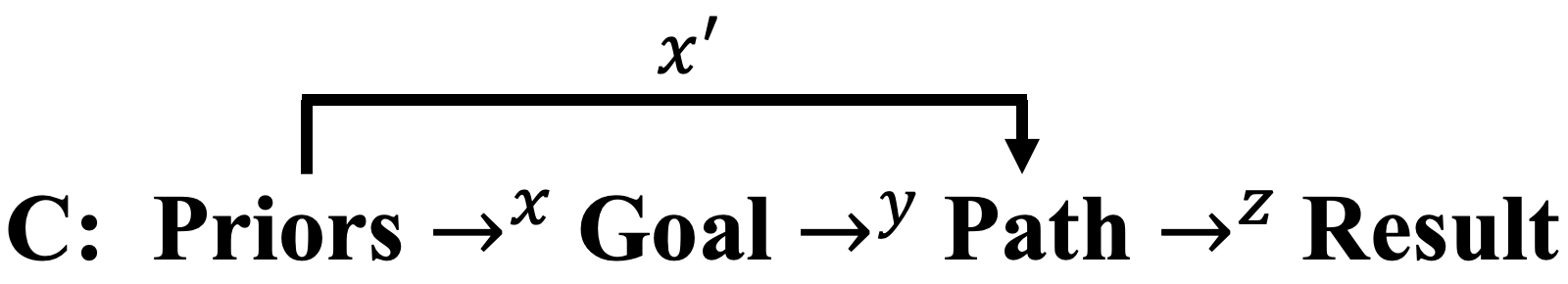}
\end{center}
\end{figure}
The main addition is that both \(\rightarrow^{x}\) and \(\rightarrow^{x^{'}}\) are conditional on
priors, knowledge and skills. In humans, necessities such as food and
shelter impose on lower-level priors (e.g., reactions to hunger and
fear; notions of causality), whereas opportunities such as higher
education and economic mobility require high-level priors (e.g., causal
understanding, associative learning, goal directedness) (\citep{cholletMeasureIntelligence2019}). The
contingencies of goals and paths on information in the forms
of priors, knowledge and skills highlight the temporal process nature
of intelligence that is central to TIS.

Temporal effects extend to the types of intelligence resources employed
throughout a lifetime. Intelligence resources in humans are characterized
by the growth of crystallized intelligence (knowledge, skills) into
adulthood (\citep{hornAgeDifferencesFluid1967}; \citep{nisbettIntelligenceNewFindings2012}) and gains in
general intelligence faculties (\citep{chaiEvolutionBrainNetwork2017}) through childhood and
adolescence. As individuals age, they may encounter fewer never-before seen
problems and are less able, for example, to maintain working memory and processing speeds (\citep{salthouseConsequencesAgeRelatedCognitive2012}). This suggests a strategy sequence in humans with a relative
shift from dependence on others (parents, social), to crystallized
(knowledge, skills) intelligence in youth, to fluid (thinking,
creativity, flexibility) intelligence in youth and mid-life, and finally more emphasis on
crystallized intelligence (with more dependence on proxies, see below)
into older ages. In other words, even if more nuanced (\citep{hartshorneWhenDoesCognitive2015}; \citep{tucker-drobCoupledCognitiveChanges2019}), evidence points to the continual
accumulation of knowledge and skills in youth enabling the ascension of
novel reasoning in early and mid-life, the latter gradually being
displaced by memory/recall into later life.

We can modify \textbf{C} to account for how feedbacks influence future intelligence resources:  	
\begin{figure}	
\begin{center}
\includegraphics[width=0.5\textwidth]{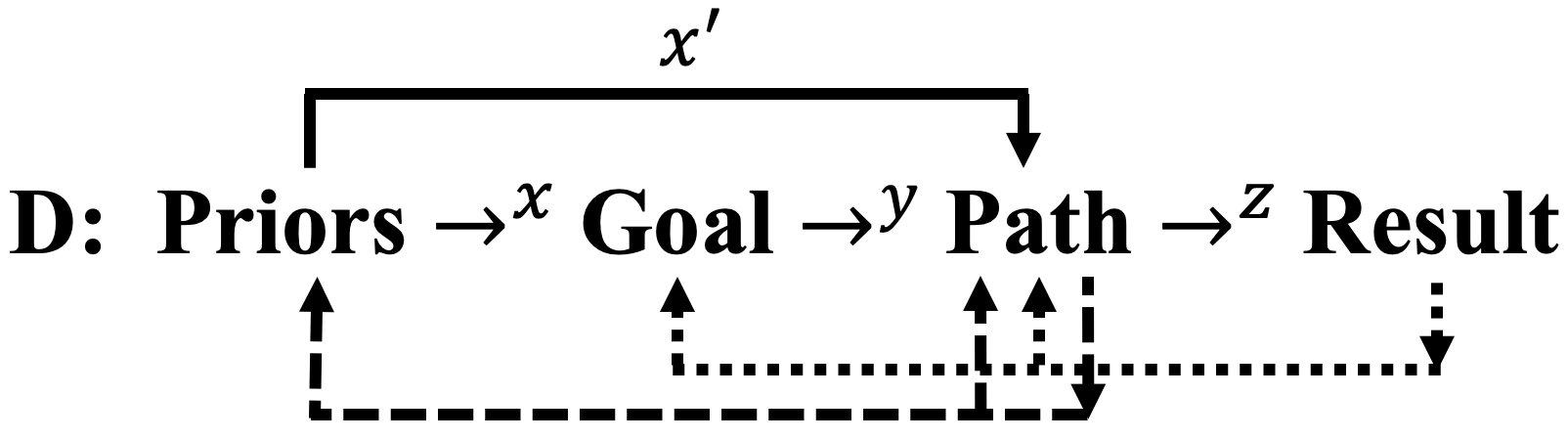}
\end{center}
\end{figure}
The dotted lines indicate how results can reinforce future paths and introduce novel set-points for future goals. Similarly, the
dashed lines indicate how path experimentation (model flexibility,
recombination of concepts) could influence future paths and
knowledge and skills. 

Even if more realistic than the sequences in
\textbf{A}-\textbf{C}, these additions are still massive oversimplifications of the sophistication of informational dynamics in biological systems. As such, \textbf{D} simply makes
the point that intelligence is a dynamic process with loops in the form
of feedbacks (comparisons) and feedforwards (predictions). The temporal (and more generally, dimensional) nature of information change is a
fundamental property of TIS and as developed below, important in describing differences in information processing among system types.

\hypertarget{Challenges and Abilities}{%
\subsection{Challenges and Abilities}
\label{Challenges and Abilities}}

The above descriptions of temporal sequences say little about system
features associated with goal definition and those promoting goal
attainment. Despite considerable discussion (e.g., \citep{spearmanGeneralIntelligenceObjectively1904}; \citep{cattellTheoryFluidCrystallized1963}; \citep{albusOutlineTheoryIntelligence1991a}; \citep{brooksIntelligenceRepresentation1991}; \citep{gershmanComputationalRationalityConverging2015}; \citep{lecunDeepLearning2015}; \citep{kabadayiDetourParadigmAnimal2018}; \citep{wangDefiningArtificialIntelligence2019}; \citep{cholletMeasureIntelligence2019}) there is no
consensus on a process-based theory of intelligence, neither for
biological divisions (\citep{baluskaHavingNoHead2016}; \citep{lyonWhatMinimalCognition2020}), nor for
artificial systems (\citep{goertzelStructureIntelligenceNew1993}; \citep{adamsMappingLandscapeHuman2012}; \citep{roliHowOrganismsCome2022}; \citep{fristonDesigningEcosystemsIntelligence2024}). This state of affairs stems in part from the lack
of agreed first principles for candidate features and the ill-defined
conceptual overlap among various traits. A way forward is a more inclusive framework based on how information is accessed and processed. I recognize that this still
falls short of a readily testable, first principles framework
and rather intend the developments below to stimulate discussion of next steps and
refinements.

The TIS framework now partitions the system into two main constructs --
challenges and abilities. Challenges are further partitioned
into (1) how agent capacities compare to goal complexity
(difficulty) or novelty (surprisal, difficulty) (\citep{sternbergTriarchicTheoryHuman1984}) and (2) the capacity to arbitrate the use of stored information
(exploitation) versus acquiring additional information (exploration)
(\citep{cohenShouldStayShould2007}; \citep{mehlhornUnpackingExplorationExploitation2015}; \citep{delgiudiceBasicFunctionalTradeoffs2018}).
Although not previously discussed in this context to my knowledge, a related
tradeoff is between (3) the employment of stored information
(crystallized intelligence) versus higher reasoning abilities (fluid
intelligence). The basis for this latter tradeoff is costs and constraints in
experiencing, learning, and accurately and efficiently storing and retrieving information in the form
of knowledge and skills, either alone, together with, or replaced to
some degree by fluid abilities such as reasoning and creativity. Thus,
for example, a computer is able to store solutions to all
possible goals and, as such, has no need for fluid abilities, but
nevertheless would need to (1) inherit these solutions or have the time
to sample them, (2) have the storage space and (3) have the processing
abilities to recall appropriate solutions when confronted with specific
goals.

Challenges are addressed based on abilities. Previous frameworks of
intelligence in humans, biology and AI have emphasized hierarchical
structure in abilities (e.g., \citep{spearmanGeneralIntelligenceObjectively1904}; \citep{carrollHumanCognitiveAbilities1993}; \citep{conradCrossscaleInformationProcessing1996}; \citep{dearyStabilityIntelligenceChildhood2014}; \citep{cholletMeasureIntelligence2019}; \citep{lecunPathAutonomousMachine2022}), but the framework presented here
is unique in distinguishing generic abilities from those actually
applied to goals and their occurrence in different system types (\hyperlink{tab1}{Table 1}). Generic abilities and goal abilities are each partitioned into multiple levels based on their
hypothetical order in evolutionary appearance, that is, the necessary establishment of
mechanisms on a given level for the emergence of more complex mechanisms
on a ``higher'' level (\hyperlink{fig4}{Figure 4}). For instance, planning manageable sequences of subgoals
serves little purpose unless there are existing capacities to solve
component tasks along a sequence. Nevertheless, solving needs references (error correction) and at higher levels the ability to arbitrate or "predict" these references. Although higher accuracy and efficiency will evolve to
some extent with solving and planning, an expectation based on evolutionary theory (\citep{smithOptimizationTheoryEvolution1978}) is that costly optimization lags
behind the inception of more grounded abilities contributing to solving and planning (e.g., reasoning, error correction, ...). Thus, the
hierarchical categorizations in \hyperlink{tab1}{Table 1} are not clear-cut (and solving and planning are
discussed in more detail in the next sections). Identifying component levels for
various abilities is thus a considerable challenge, both because of the
subjective nature of their definition and since abilities may articulate
at more than one level.

\hyperlink{tab1}{Table 1} also lists system types commonly associated with different
abilities. Exceptions exist, notably in biology, where for example, certain avian
lineages and cephalopods match or even surpass some mammals
in generic and goal-associated abilities (\citep{emeryMentalityCrowsConvergent2004}; \citep{edelmanAnimalConsciousnessSynthetic2009}; \citep{amodioGrowSmartYoung2019}). There are, nevertheless, regularities that characterize intelligence in
the major system classes discussed here. Perhaps the most controversial of these are physio-chemical systems. In physio-chemical systems, events depend on the
actions and influences of laws and environments. Physical systems do
not have ``abilities'' as such, yet they do rearrange information based on
thermodynamic laws, possibly resulting in increased information
relative to initial conditions (e.g., randomly distributed minerals to
arranged crystal lattices). Thus, at least abstractly, physical systems can reactively resolve uncertainty.

\hypertarget{MODELLING THE THEORY OF INTELLIGENCES}{%
\section{MODELLING THE THEORY OF INTELLIGENCES}
\label{MODELLING THE THEORY OF INTELLIGENCES}}

Theoretical developments to characterize intelligence are found among
many disciplines and diverse systems (e.g., \citep{sternbergTriarchicTheoryHuman1984}; \citep{albusOutlineTheoryIntelligence1991a}; \citep{goertzelStructureIntelligenceNew1993}; \citep{wangAbstractIntelligenceUnifying2009}; \citep{peraza-vazquezBioInspiredMethodMathematical2021}). Explicit models have described factors
such as temporal sequences from access to priors and accumulating
knowledge and skills through to learning (\citep{cholletMeasureIntelligence2019}), and prediction
error-checking and correction in achieving goals (\citep{lecunDeepLearning2015}; \citep{lecunPathAutonomousMachine2022}).
These and other recent advances have largely centered on AI, both
because of the rapid growth in artificial intelligence and since these
systems are more tractable than biological intelligence (and human
intelligence in particular). The macroscopic models presented below are sufficiently
flexible to apply to a range of intelligence systems and contexts. The models
accommodate possibilities that goals can be attained with little or no
insight, or that intellectual capacities do not ensure goal realization
(e.g., \citep{mashburnWorkingMemoryIntelligence2023}).

The framework developed so far sets the stage for theory
incorporating temporal, multi-level and multi-scale factors, focusing on
how intelligence is a response to uncertainty and complexity. I hypothesize that at a macroscopic level, intelligence in humans (and
possibly other select animal species) operates at local (node, subgoal) and
regional or global (network, modular) levels. Building on the discussion in the
previous section, local \textit{solving} is addressing an arbitrarily small (manageable or imposed) unit of what might be a global objective, an example being single
moves in the game of chess. As a goal becomes increasingly complex, multiple
abilities may be marshalled in achieving solving, but this may be insufficient to address higher-dimension goals and particularly those where uncertainties themselves unpredictably change (e.g., an opponent in chess makes an unexpected move). 

Uncertainty in achieving high-dimensional goals is reduced
through \textit{planning}. Planning -- which appears to be a uniquely highly
developed characteristic of humans (\citep{sternbergTriarchicTheoryHuman1984}; \citep{albusOutlineTheoryIntelligence1991a}, but see \citep{rothEvolutionBrainIntelligence2005}; \citep{smithAnimalMetacognitionTale2014}; \citep{roliHowOrganismsCome2022}) -- can involve
(as does solving) one or more of the non-mutually exclusive capacities
of computation, abstract reasoning, mechanistic/causal understanding,
and creativity (\hyperlink{tab1}{Table 1}). Planning is central to optimization,
that is, path efficiency and goal accuracy and completeness. Optimal path trajectories to
goals become more difficult to attain as the goals themselves become
more complex. Non-planned, stepwise, myopic strategies will tend to
decrease the predictability of future path nodes and therefore accurate
and efficient goal outcomes (\citep{kauffmanInvestigations2000}). Thus, planning tends to optimize
information use and foresee and dynamically adapt to uncertainty,
thereby reducing surprises and achieving more efficient resolutions
(\citep{maierUncertainFutureDeep2016}; \citep{walkerAdaptPerishReview2013}). Planning can also reduce
goal difficulty by engineering the surrounding environment or altering
goals themselves on-route, meaning, for example, that satisficing as
opposed to optimization may emerge as the most intelligent
outcome (\citep{simonRationalChoiceStructure1956}; \citep{hayes-rothSatisficingCycleRealtime1994}; \citep{bossaertsComputationalComplexityHuman2017}).

In the models below, solving and planning are each represented by a key variable. For solving
it is \emph{U\textsubscript{n}}, the \emph{useful} information acquired
at subgoal \emph{n} towards the ultimate goal, whereas for planning it is
\emph{A\textsubscript{n}}, the accuracy, efficiency and completeness of the informational path traced to the ultimate
goal. Nodes (units) form a sequence of subgoals that increasingly approach
the ultimate goal if the system better solves and for complex goal, plans the sequence. The net change in information between start and finish of a
sequence could be either positive (some degree of success) or negative (failure), yet intelligence only manifests if information is at least transiently gained at one or more nodes.

\hypertarget{Solving}{%
\subsection{Solving}\label{Solving}}

We assume a goal can be represented by a network of nodes, each node
corresponding to an information state (subgoal) relative to the ultimate
objective. The agent engages in the network at \emph{n}=1 with stored
information in the form of priors, knowledge and skills and thereafter
uses this together with external data through chosen path nodes towards no, partial, or complete
goal resolution at node \emph{n}=\emph{N}\(\geq\) 1. We assume uncertainty (information) is
fixed at each node, that is, forging different paths through any given
subgoal does not change the information content of that
subgoal.

The solving component of intelligence \(\mathbb{U}_{y}\) is the gain in information useful for goal \emph{y}, normalized by the number of subgoals \emph{N}

\hypertarget{1a}{}
\large
\(\mathbb{U}_{y}\  = \ \cfrac{1}{N}\ \mathlarger{\mathlarger{{\sum_{n = 1}^{N}}}} \ {U_{n,y} \{ \cdot\} }  / {{\hat{U}}_{n,y}}\) \ \ (1a)

\normalsize

where \(U_{n,y}\) and \({\hat{U}}_{n,y}\) are, respectively, the realized gain in information (assumed equivalent to the reduction in uncertainty)
at node \emph{n} for goal \textit{y}, and the amount
of goal-related information available. \(U_{n,y}\{\cdot\}\) will
be a function of agent ability, subgoal uncertainty, disruptive noise, and surrounding environmental conditions. Disruptive noise and inhospitable environments could in principle result in information loss (time course ii in \hyperlink{fig1}{Figure 1B}). Whereas \({{U}}_{n,y}\) will be a complex function, the sequence of \({\hat{U}}_{n,y}\) is given (i.e., there is no planning; this will be relaxed below). By
definition \({\hat{U}}_{n,y}\) \(\geq\) \(U_{n,y}\) and \(\mathbb{U}_{y}\) approaches
1 as \(U_{n,y}\) → \({\hat{U}}_{n,y}\) for all \emph{n}. \(\mathbb{U}_{y}\) is normalized by the number of subgoals \textit{N}, which does not explicitly account for solving efficiency. This could be incorporated as a first approximation by dividing \(U_{n,y}\) by the time elapsed on each subgoal \textit{n}. (Evidently, there will be a tradeoff: overly rapid solving will result in lower information gained). Note that equation (1a) is a compact macroscopic form for a diverse set of implicit micro-
and mesoscopic processes and although not developed here, can be related to information theory (e.g., \citep{adamiWhatComplexity2002}), and more specifically entropy (\({\hat{U}}_{n,y}\)) and untapped information (\({\hat{U}}_{n,y}\) -- \(U_{n,y}\)), the latter which reflects difficulty.

Importantly, eqn (1a) makes no assumption as to whether or to what extent
the goal is attained, that is, even should all visited nodes be
completely resolved it is possible that the ensemble of information is
goal-incomplete. Rather, (1a) only quantifies the resolution of
uncertainty for chosen nodes under the implicit assumption of local, non-planned node choice. 

We can modify (1a) to include the contribution
\({R}_{n,y}\) of information at node \emph{n} to
goal completion, with the condition \(\sum_{n = 1}^{N} {R_{n,y}/{N}\ \leq \ 1}\):

\hypertarget{1b}{}
\large
\(\mathbb{U}_{y}^{R}\  = \ \dfrac{1}{N}\ \mathlarger{\mathlarger{{\sum_{n = 1}^{N}}}} \ R_{n,y} {U_{n,y} \{ \cdot\} } / {{\hat{U}}_{n,y}} \) \ \ (1b)
\normalsize

Thus, \(\mathbb{U}_{y}^{R} \rightarrow 1\ \) as both uncertainty
reduction is maximal at each node \textit{and} the sum of information gained at
all nodes is accurate and goal complete. 

Goals can be attained through some combination of solving
\(U_{n,y}\), insight in nodes selected \(R_{n,y}\) and number of nodes
visited \emph{N}. Importantly, \(R_{n,y}\) \textit{does not} assume long-range
planning, but rather reflects the quality of adjacent node choice. In not being normalized by nodes \textit{N} or time, the numerator in eqn (1b), \(\mathbb{X}=\sum_{n = 1}^{N}{R_{n,y} U_{n,y}}\) is a measure of realized goal complexity. Note that \(\mathbb{X}\) can also be interpreted as a stopping condition, either an arbitrary goal-accuracy threshold (\(\mathbb{X}\)\textgreater{}\(\mathbb{W}\))
or a margin beyond a threshold (\(\mathbb{X}\)-\(\mathbb{W}\)>0). An example of a
threshold is the game of chess where each move tends to reduce
alternative games towards the ultimate goal of checkmate, but a
brilliant move or sequence of moves does not necessarily result in victory. Examples
of margins are profits in investments or points in sporting events,
where there is both a victory threshold (gain vs loss, winning vs
losing) and quantity beyond the threshold (profit or winning margin). We
do not distinguish matching from maximization below, but rather
highlight that goal resolution could involve one or both.

\hypertarget{Planning}{%
\subsection{Planning}\label{Planning}}

Solving does not necessarily ensure goal attainment
(0<\(\mathbb{X}\)\textless{}\(\mathbb{W}\)). This is because, for example, a
system may either (1) correctly execute most of a complex series of
computations but introduce an error that results in an inaccurate or incomplete
resolution, or (2) choose nodes that, taken together, only contain a
subset of the information necessary to accurately and completely attain the goal.

We begin by modelling hierarchical planning \(\mathbb{A}_{y}\) on goal \textit{y} as the goal-related
information contained in the actual choices \({\hat{U}}_{h,n,y}\) of path nodes over node sequence \textit{n}=1 to \({{N}}\{h\}\) (the local planning horizon) relative to the optimal sequence \({\tilde{U}}_{h,g,y}\) (i.e., that maximizes information) summed over \textit{g}=1 to \({{G}}\{h\}\) 

\hypertarget{2a}{}
\large
\(\mathbb{A}_{y} = \ \mathlarger{\mathlarger{{\sum_{h = 1}^{H}}}} \ \biggr(\mathlarger{\mathlarger{{\sum_{n = 1}^{N\{h\}}}}} {\hat{U}}_{h,n,y}\ / \mathlarger{\mathlarger{{\sum_{g = 1}^{G\{h\}}}}} {\tilde{U}}_{h,g,y} \biggr) \) \ \ (2a)
\normalsize

where the sum of the terms in parentheses is itself summed over \textit{H} sequences. The capacity to plan therefore partitions the goal into local \textit{n} and higher-level \textit{h} planning. In the lower limit of \textit{H}=1, all planning occurs when the goal is engaged at node \textit{n}=1. For \textit{H}>1, planning occurs in multiple sequential sequences (i.e., the global planning horizon). Importantly, the optimal sequence could involve a different number of nodes \textit{G}\(\neq\)\textit{N}, but optimal information will always be greater than or equal to information in the sequence actually planned (the term in parentheses is less than or equal to 1).

Both actual and optimal local planning (\({\hat{U}}_{y}\), \({\tilde{U}}_{y}\)) and their horizons (\textit{N}, \textit{G}) are a function of the global planning sequence, \textit{h}. Furthermore, both numerator and denominator of (2a) can be functions of potential information (\({\hat{U}}_{y}\)) and/or acquired information (\(U_{y}\)) from the past (1..\textit{h}-1), present (\textit{h}) and/or expected in the future (\textit{h}+1 ...). Expression (2a) -- despite its complexity -- does not model how node choices are actually made, nor does it account for path efficiency (i.e., number of nodes taken; spent time or energy).  

Equation (2a) can be greatly simplified by assuming that planning occurs only once when the goal is engaged:

\hypertarget{2b}{}
\large
\(\mathbb{A}_{y}^{G} = \ \mathlarger{\mathlarger{{\sum_{n = 1}^{N}}}}{{\hat{U}}_{n,y}/N} \ \mathlarger{\mathlarger{{/}}} \ \mathlarger{\mathlarger{{\sum_{g = 1}^{G}}}}{\tilde{U}}_{g,y}/G \) \ \ (2b)
\normalsize

where the numerator now accounts for efficiency as the normalized sum over the \emph{N} nodes and
the denominator is the normalized sum over the number \emph{G} of nodes potentially
leading to the accurate and complete resolution of the goal. Note that one or more paths might result in acquiring full goal information
\({\tilde{U}}_{g,y}\) and we assume for simplicity that 
the denominator is minimized (i.e., optimal planning). 

A final index for goal attainment \(\mathbb{A}_{y}^{S}\) quantifies \textit{solving} based no planning (eqn (1b)), relative to that based on optimal planning

\hypertarget{2c}{}
\large
\(\mathbb{A}_{y}^{S} = \ \mathbb{U}_{y}^{R}\ /\ \mathbb{U}_{y}^{G}\) \ \ (2c)
\normalsize

where
\(\mathbb{U}_{y}^{G}=\sum_{g = 1}^{G}{U_{g,y}}/{{\tilde{U}}_{g,y}} {G}\) and it is assumed that optimal planning is complete, i.e., \(\sum_{n = 1}^{G}{R_{g,y}/{G}\ =1}\). 

\hypertarget{Intelligences}{%
\subsection{Intelligences}\label{Intelligences}}

Any multifactorial definition of intelligence is frustrated by the
arbitrariness of component weighting. In this respect, a central
ambiguity is the relative importances of marshalling priors, inventing,
choosing or being confronted with a goal, articulating the path to the
goal, and achieving the desired result. Clearly, each phase depends to
some extent those preceding it and each phase may also depend on
predictions of those phases not having yet occurred. For illustration of
the basic issue, consider two different algorithms available to an
entity -- an efficient (smart) one and an inefficient (dumb) one -- each
yielding an answer to the same problem. There are two outcomes for each
algorithm. Dumb algorithm, wrong answer (0,0); Smart algorithm, but
wrong answer (1,0); Smart, correct (1,1); And yes, especially for
multiple choice questions, Dumb, correct (0,1). Undoubtedly, (1,1) and
(0,0) are the maximal and minimal scores respectively. But what about
(1,0)'s rank compared to (0,1)? If a multiple-choice test, then one gets
full points for what may be a random guess and (0,1). If the path taken
is judged much more important than the answer, then an interrogator
would be more impressed by (1,0) than (0,1).

\(\mathbb{U}\) and \(\mathbb{A}\) are indicators of two complementary
forms of intelligence in those systems possessing both solving and planning. Although they can each serve as stand-alone assessments, they are ultimately interrelated by
\({\hat{U}}_{n}\), information content at each node
visited (see below). Correlations between \(\mathbb{U}\) and
\(\mathbb{A}\) will be promoted by goal simplicity, that is either few
possible nodes to accurate and complete resolution, or many nodes, but simplicity in
their uncertainties and in the path to accurate resolution. We do not
explicitly model these co-dependencies and rather recognize that goal
complexity will tend to lower associations between abilities to solve and plan.

\hyperlink{fig5}{Figure 5} shows how an agent's \(\mathbb{A}_{y}\)--\(\mathbb{U}_{y}\)
trajectory hypothetically plots onto a space of achievements. Different
nodal paths can produce the similar final outcome (S-C and S-O), and very similar paths can lead to different outcomes (S-B and S-C). Underscoring the importance of flexibility
in attaining goals, one or the other of \(\mathbb{U}_{y}\ \)and
\(\mathbb{A}_{y}\) might increase, decrease or remain unchanged as the
agent proceeds through successive nodes. This underscores the expectation that the most efficient and accurate path to goal completion will involve the optimal allocation of (possibly competing) solving and planning abilities (\textit{cf} trajectories i or ii vs iii in \hyperlink{fig1}{Figure 1A}). \hyperlink{fig5}{Figure 5} also illustrates that
solving and planning need not concord (e.g., path S-B
in \hyperlink{fig5}{Figure 5}). Nevertheless, given likely correlations among intelligence
components, we predict that there will be a feasible parameter space for a given
agent's abilities (shaded area, \hyperlink{fig5}{Figure 5}).

\hypertarget{A General Equation}{%
\subsection{A General Equation}\label{A General Equation}}

Expressions (1a-b) and (2a-c) provide a macroscopic basis for a theory
of intelligence. They characterize two central abilities, solving and planning,
but as related above, overall goal performance can (but does not necessarily) involve both capacities.
A more general index for intelligence is based on the assumption that solving is necessary and
planning augments solving capacities. Based on equations (1b) and (2b), we have a simple
expression for the intelligence of an agent with capacities \emph{x} addressing goal
\emph{y}

\hypertarget{3a}{}
\large
\(\mathbb{I}_{x,y} = \mathbb{U}_{x,y}^{R}\ {({\alpha} + {\beta}\mathbb{A}}_{x,y}^{G})\) \ \ (3a)
\normalsize

where \(\alpha\) and \(\beta\) a positive constants and \(\alpha\) + \(\beta\) = 1. Planning becomes increasingly necessary to intelligence as \(\beta \rightarrow\) 1. Equation (3a) is an expression of both goal attainment \(\mathbb{U}_{y}^{R} \) and the quality of the route taken \(\mathbb{A}_{x,y}^{G}\). Substituting information terms from eqns (1b) and (2b) and dropping subscripts and summations, we have

\hypertarget{3b}{}
\large
\(\mathbb{I}_{x,y} = {R U} (\dfrac{\alpha}{{\hat{U}}} + \dfrac{\beta}{{{\tilde{U}}}} )\) \ \ (3b)
\normalsize

Intelligence is the information actually gained as a fraction of the summed, weighted inverse entropies of the actual and optimal path. Note that when the agent takes the optimal path (i.e., \textit{R}=1, \({\hat{U}}\)=\({\tilde{U}}\)), (3b) reduces to 

\hypertarget{3c}{}
\large
\(\mathbb{I}_{x,y} = {{U}} \ / \ {{{\tilde{U}}}}\) \ \ (3c)
\normalsize

where \textit{U}/\({\tilde{U}}\) is now summed from 1 to \textit{G}. 

Although \(\mathbb{I}_{x,y}\) is relative to potential information or entropy (denominators of (3b)), an additional step is necessary to relate intelligence to difficulty. Thus, intelligence defined as \textit{achievement relative to difficulty} can be
quantified by first proposing an expression for difficulty based on
concepts developed above

\hypertarget{3d}{}
\large
\(\mathbb{D}_{y} = {\mathbb{C}_{g,y}} \ / {\mathbb{\ {Q}}_{x,y}}\) \ \
(3d)
\normalsize

\(\mathbb{C}_{g,y} = \sum_{g = 1}^{G}{\tilde{U}}_{g}\) is the intrinsic complexity of goal \emph{y}, that is, the minimum amount of information required to accurately and completely achieve (or describe) the goal. \(\mathbb{{Q}}_{x,y}\) > 0 is the
expected ability of the agent with expertise \emph{x} to achieve goals in class
\emph{y}. All else being equal, \(\mathbb{{Q}}_{x,y}\) will decrease as \emph{x} and \emph{y} diverge (i.e., greater
surprisal). Because actual experience on a goal reduces future surprisal on similar goals, \(\mathbb{{Q}}_{x,y}\) will be challenging if not impossible to estimate for a given agent, necessitating an index based on benchmarked ability (see below). 
As ability exceeds goal complexity \(\mathbb{D}_{y} < 1\) the goal becomes increasingly simple, whereas the inverse \(\mathbb{D}_{y} > 1\) indicates increasing goal difficulty. 

Whereas \(\mathbb{I}_{x,y}\) from equations (3a-c) is a useful measure of intelligence based on actual solving and planning, a complementary index for system \textit{x} on goal \textit{y} based on the notion of difficulty is given by

\hypertarget{3e}{}
\large
\(\mathbb{\hat{I}}_{x,y} = \mathbb{D}_{y} \ {\mathbb{I}_{x,y}} \ / \ {\mathbb{Q}_{x,y}}\ \)  \ \ (3e)
\normalsize

High intelligence implies a high performance on a difficult goal \(\mathbb{C \gg Q \approx I}\), whereas \(\mathbb{C \ll Q \gg I}\) signals lower than expected performance on an easy goal. 

Equation (3e) is both general and powerful. \(\mathbb{D, I}\) and \(\mathbb{Q}\) are all composed of terms in units of information. By simple rearrangement of difficulty (eqn (3d)) we have an expression for the expectation of intelligence for a given level of complexity \(\mathbb{Q / C}\). The mean and variation in \(\mathbb{I / Q}\) over many similar tasks is a measure of positive impacts of training (AI) or experience (biology), or negative impacts of environmental variation or noise. Moreover, \(\mathbb{Q}\) could be subdivided into ability that is transmitted during a lifetime, and ability that is evolutionarily embodied in system phenotype. Finally, difficulty in (3e) can be cast as performance relative to a population benchmark \(\mathbb{\Bar{Q}}_{bench,y}\)

\hypertarget{3f}{}
\large
\(\mathbb{\hat{I}}_{bench} = {\mathbb{C}_{g,y} \ \mathbb{I}_{x,y}} \ / \  {\mathbb{\Bar{Q}}_{bench,y}^2}\)  \ \ (3f)
\normalsize

\hypertarget{Transmission and Evolution}{%
\subsection{Transmission and Evolution}\label{Transmission and Evolution}}

The above mathematical developments characterize intelligence
as the resolution of subgoals -- and possibly the resolution of a path -- culminating in a reactive result or a defined goal. The equations focus on a single system addressing a single goal and as such are unable to capture larger scales of intelligence, notably the \emph{transmission} of capacities or traits promoting intelligence, and their \emph{evolution}. 

Transmission can be
an important facilitator of intelligence and come from diverse sources.
Examples include social interactions (e.g., collectives; \citep{krauseSwarmIntelligenceAnimals2010}) and technology (e.g., tools; \citep{biroToolUseAdaptation2013}). Social
interactions are particularly interesting since they can range from an
individual benefiting from or depending on one or
more others (e.g., outsourcing), to a larger group
where individuals exchange information and act collectively towards a goal (e.g.,
certain microbial or animal aggregations; \citep{couzinCollectiveCognitionAnimal2009}), to a division of
intelligence problem solving where different functions in goal attainment are
distributed among individuals (e.g., social insects, social mammals,
human corporations; \citep{pacalaEffectsSocialGroup1996}). The intelligence substrates provided by these and
other ``proxies'' could complement, substitute, enhance or extend an
autonomous individual system's own facilities (\citep{leeOutsourcingMemoryNiche2022}). For
example, in human social learning, resolving tasks or goals is the raw
material for others to observe, record and emulate, thereby contributing
to the diffusion and cumulative evolution of knowledge (\citep{roliHowOrganismsCome2022}; \citep{miglianoOriginsHumanCumulative2022}; \citep{henrichWhatMakesUs2023}).
In embodying knowledge, skills and their transmission, culture and
society are at the foundation of intelligence in humans and many other animal
species.

We can incorporate proxies into equation (3e) in a simple way by assuming they reduce goal
difficulty, whereby 

\hypertarget{4}{}
\large
\(\mathbb{\hat{I}_P} = \dfrac{\mathbb{C}_{g,y}}{(h\mathbb{P}_{y} + \mathbb{Q}_{x,y})}\ \dfrac{\mathbb{I}_{x,y}\{\mathbb{P}_{y}\}}{\mathbb{Q}_{x,y}}\ \)  \ \ (4)
\normalsize

where \textit{h}\(\mathbb{P}_{y}\) is reduction in difficulty given the action of the proxy, and the agent's capacity is modified (increased) as per the undefined function \(\mathbb{I}_{x,y}\{\mathbb{P}_{y}\}\). Equation (4) is a compact description of how proxies influence both goal difficulty (with constant \textit{h}) and realized performance \(\mathbb{I}_{x,y}\{\mathbb{P}_{y}\}\). Thus, proxies decrease intelligence by decreasing difficulty, but increase intelligence by increasing performance. The balance of these two effects determine the net impact of proxies on intelligence as defined in eqn (4).

Two other ways in which intelligence capacities can be transmitted are
genetically or culturally. To the extent that
inherited traits impinging on intelligence follow the same evolutionary principles as
other phenotypic traits, one should expect that some or all of genetically influenced, intelligence traits are labile and evolve, leading to alterations in existing traits
or the emergence of novel traits. Despite contentious discussion of
heritable influences on intelligence in humans (\citep{gouldMismeasureMan1981}; \citep{herrnsteinBellCurveIntelligence1994}), the more general question of the biological
evolution of intelligence continues to receive dedicated attention (\citep{chiappeEvolutionDomainGeneralMechanisms2005}; \citep{flinnEcologicalDominanceSocial2005}; \citep{rothEvolutionBrainIntelligence2005}; \citep{lazerNetworkStructureExploration2007}; \citep{wangAbstractIntelligenceUnifying2009}; \citep{sterelnySocialIntelligenceHuman2007}; \citep{gaboraEvolutionIntelligence2011}; \citep{pinkerCognitiveNicheCoevolution2010}; \citep{burkartEvolutionGeneralIntelligence2017}). That intelligence traits evolve and present
patterns consistent with evolutionary theory is evidenced by their
influence on reproductive fitness (\citep{burkartEvolutionGeneralIntelligence2017}) including
assortative mating (\citep{plominGeneticsIntelligenceDifferences2015}), and age-dependent
(senescent) declines in fluid intelligence (\citep{craikCognitionLifespanMechanisms2006}).
Nevertheless, one of the main challenges to a theory of the evolution of
intelligence is identifying transmissible genetic or cultural variants
that contribute to intelligence ``traits''. Despite limited knowledge of
actual gene functions that correlate with measures of intelligence (\citep{sabbChallengesPhenotypeDefinition2009}; \citep{savageGenomewideAssociationMetaanalysis2018}), a reasonable hypothesis is that
intelligence is manifested not only as active engagement, but also in phenotypes
themselves, that is, evolution by natural selection \emph{embodies}
intelligence in functional entities (e.g., proteins, cells, organ
systems\ldots) at different biological levels (see also \citep{adamiWhatComplexity2002}; \citep{shreeshaCellularCompetencyDevelopment2023}).

The detailed development of how TIS relates to the evolution of intelligence is beyond the present study. Nevertheless, as a start, we expect that the change in the mean level of intelligence trait \textit{x} in a population will be proportional to the force of selection on task \textit{y}. Simplifying this process considerably we have

\hypertarget{5}{}
\large
\(\Delta{\Bar{x}} \ \propto \ {p}_{y} \ {\mathbb{\hat{I}}_{x,y}} / \Bar{\mathbb{I}}_{y} \) \ \ (5)
\normalsize

where \({p}_{y}\leq1\) is the proportional contribution of task \textit{y} to mean population fitness. The response to selection will be expressed in changes in trait \textit{x} that affect benchmarked expected ability (i.e., the benchmark \({\mathbb{Q}_{x,y}}\) evolves) and realized performance of individuals harboring the changed trait \({\mathbb{\hat{I}}}_{x,y}\) (eqn (3b)) relative to the population mean \(\mathbb{\bar{I}}_{y}\), these latter two quantities assumed to be proportional to fitness. If task \textit{y} has minor fitness consequences (either \textit{y} is rarely encountered and/or of little effect when encountered), then selection on trait \textit{x} will be low. Note that equation (5) can be generalized to one or more traits that impinge on one or more tasks (\citep{arnoldMorphologyPerformanceFitness1983}). Moreover, this simple expression implies that difficult, novel encounters have larger relative payoffs and thus, insofar as choices are made to accomplish goals of a given difficulty, fitness is proportional to the achievement level \({\mathbb{I/Q}}\). 
\\

The full framework for TIS has a four-part
structure (\hyperlink{fig6}{Figure 6}), now including transmission within an individual
system's lifetime (information input, perception, learning) and
evolution in populations (cultural, genetic). A given system's
intelligence is thus the integration of (I) the core system (controller, processor, memory), (II) extension to proxies (niche
construction, social interactions, technologies), (III) nascent,
communicated or acquired abilities (\hyperlink{tab1}{Table 1}) and information (priors,
knowledge, skills), and (IV) traits selected
over evolutionary time. Equations (1-3) only cover point (I) and only on
a macroscopic scale. Equation (4) incorporates how proxies (point (II))
complement intrinsic system capacities, but not how they are
actively used in solving and planning. Equation (5) is a start to account for evolutionary changes in intelligence traits (point (IV)).

\hypertarget{DISCUSSION}{%
\section{DISCUSSION}\label{DISCUSSION}}

Until relatively recently the study of intelligence focused on
psychometrics in humans. With the development of AI and a greater
emphasis on interdisciplinarity over the past few decades, progress has
been made towards theory that spans humans and other systems, with the
prospect of a general theory of intelligence based on first principles. Here, I survey
recent advances towards conceptual unification of definitions of
intelligence, arguing that a framework needs to incorporate
\emph{dimensions} (levels and scales) and be applicable to any
phenomenon or system fitting the general definition of \textit{uncertainty
reduction producing an intentional goal (active) or unintentional result (reactive)}. Levels include
capacities to perceive, sort, and recombine data, the employment of
higher-level reasoning, abstraction and creativity, the application of
capacities in solving and planning, and optimizing accuracy and
efficiency. Scales enter in diverse ways, including but not limited to
how intelligence becomes evolutionarily embodied in system phenotypes,
how abilities are acquired or lost over the life of a system, and the more usual descriptor of intelligence: how
capacities are actively applied to defining and achieving goals. The
theory I propose -- the Theory of Intelligences (TIS) --
consists of a framework (\hyperlink{fig6}{Figure 6}) incorporating key
levels and scales (\hyperlink{tab1}{Table 1}), and mathematical models of intelligence characterizing
local scale solving and larger scale planning (equations 1 -- 4). Thus, TIS is parsimonious in ignoring the microscopic detail that underlies the core phenomena of solving, planning and optimization. TIS contrasts notably with frameworks in which intelligence is partitioned into reflex/reaction and planning (\citep{mattarPlanningBrain2022}; \citep{lecunPathAutonomousMachine2022}). Specifically, in LeCun's Mode 1, the system reacts to perception, perhaps minimally employing reasoning abilities. In Mode 2, the system can use higher reasoning abilities to plan, possibly in a hierarchical fashion. TIS differs from this framework by conceptualizing reflex/reaction \emph{as part} of both solving and planning, and rather distinguishes subgoal phenomena from abilities to identify two or more subgoals, based on their accuracy and completeness relative to an intentional objective or unintentional result. A priority for future work is to relate how generic abilities (e.g., error correction, reasoning) map onto the goal-specific abilities of solving, planning and their optimization. 

TIS is grounded in the idea that intelligence is an operator that
differentiates, correlates and integrates data and information,
rendering entropy into new information. Depending on the goal, an outcome of
intelligence differs from the data on which it was derived and may involve any form, from random to heterogeneous, to uniformly
structured, varying or unvarying in time, distributed in space, etc.
Some goals, such as checkmate in chess resolve uncertainty in each move
(solving subgoals), can anticipate future moves (planning across subgoals), and occur in huge but
defined spaces leading to one of three finishes (checkmate,
resign or draw). Other endeavors such as art occur in subjective spaces
and produce novel form (sculpture) or novel routes to a finality (cinema).
Yet other objectives such as military or political ones may have
significant degrees of uncertainty, occluded information, and
unpredictable outcomes. These general examples support the idea that outcomes
and the means to attain them are characteristic of different goal types.
All have in common information transformation and some involve creation; that is,
the differences and uncertainty in starting conditions relative to
the path and/or the structure of the outcome. Interestingly, the above arguments parallel McShea's
(\citep{mcsheaPerspectiveMetazoanComplexity1996}) classification of types of complexity, whereby complexity can be
in one or more of process, outcome and/or the levels or hierarchies in
each. TIS generalizes previous theoretical treatments (e.g.,
\citep{cholletMeasureIntelligence2019}; \citep{lecunDeepLearning2015}; \citep{lecunPathAutonomousMachine2022}), fostering their interpretation in information theory and the ecological and evolutionary sciences.

TIS posits intelligence as a universal operator of uncertainty reduction
producing organizational change. This has interesting implications.
First, natural and artificial selection are forms of intelligence, since
they reduce uncertainty and increase information (e.g., \citep{frankNaturalSelectionHow2012}).
Inspired by life history theory (\citep{stearnsLifeHistoryEvolution2000}; \citep{kaplanTheoryHumanLife2000}),
selection is expected to act on heritable traits that can affect accuracy
and efficiency in existing intelligence niches, and those that affect
flexibility, generalization and creativity in new intelligence niches. These
``either/or'' expectations of trait evolution will depend to some extent
on contexts and tradeoffs (\citep{delgiudiceBasicFunctionalTradeoffs2018}). Second, organizational change could involve some combination of assembly,
disassembly, rearrangement, morphing and recombination. It is an open
question as to how information dynamics via these and
other modes affects complexity (\citep{adamiWhatComplexity2002}; \citep{wongRolesFunctionSelection2023}).
According to TIS, intelligence decreases uncertainty, but it can also transiently increase it (either due to error or counter-intuitive but optimal pathways), and intelligence can
either increase, decrease or leave unchanged final complexity. These
observations do not contradict the prediction that intelligence is
associated with increased information, \emph{either} in the
path to a goal \emph{and/or} in the goal itself (see also \citep{adamiWhatComplexity2002}).
Together with the first implication (natural selection) this is consistent with the idea
that evolutionary selection embodies intelligence into (multi-scale and multi-level)
structures and functions. Thus, memory -- a key component of intelligence -- can be
stored in one or more ways: inherited as phenotypes, stored
electrochemically as short- or long-term memory,
or stored in proxies such as society, collectives, technology and
constructed niches (\hyperlink{fig6}{Figure 6}). The extent to which phenotypes and
proxies are a compendium of ``information past'' and
``intelligences past'' is an open question.
\\

\hypertarget{Predictions}{%
\subsection{Predictions}\label{Predictions}}

TIS makes a number of testable predictions. All are interrelated, which is both a strength and a weakness of the theory. One of the central concepts is the asymmetric interdependence between solving and planning, and how this addresses variation in complexity.

SOLVING, PLANNING AND COMPLEXITY. The interrelations between intelligence and complexity are at the foundation of TIS. Specifically, intelligence traits are expected to broadly concord with the complexity of the goals relevant to the fitness of the system, i.e., the intelligence niche (\hyperlink{fig2}{Figure 2}). The intelligence niches of any two individuals of the same system type will overlap in some dimensions (e.g., basic food preferences), but possibly not in others (ability to play a musical instrument). Capacities are expected to have general characteristics across niches, but also specificity in certain niches. Solving is required for goals of any complexity, whereas only as goals become sufficiently complex relative to the abilities of the system does planning (which minimally entails costs in time and energy) become useful and even necessary (\hyperlink{fig4}{Figure 4}). TIS predicts goal achievement will correlate with solving and planning, correlating more strongly with capacities of the former over sufficiently low complexities, and more strongly with abilities of the latter over sufficiently high complexities (\hyperlink{fig1}{Figure 1A}, \hyperlink{fig4}{Figure 4}). Although I know of no evaluation of this prediction, the results of Duncan and colleagues (\citep{duncanComplexityCompositionalityFluid2017b}) showed that proxy planning (an examiner separating complex problems into multiple parts) equalized performance among human subjects of different abilities.

PROXIES. Proxies are an underappreciated and understudied influence on intelligence. Tools and technology evolve both in terms of efficiency and diversity (\citep{hochbergInnovationEmergingFocus2017}). Despite this, the specific roles of proxies in system intelligence remain largely unknown. One prediction of TIS is proxies such as tools, transportation and AI can function to help solve local, myopic challenges, and in augmenting the system’s own ability to solve, make planning logistically feasible. Proxies that themselves are able to plan can be important in achieving complex goals out of the reach of individual systems (e.g., human interventions à la \citep{duncanComplexityCompositionalityFluid2017b}). Human social interactions have served this function for millennia and the eventual emergence of AGI could assume part of this capacity (\citep{lecunPathAutonomousMachine2022}), possibly as proxy collectives (\citep{duenez-guzmanSocialPathHumanlike2023}). TIS also predicts that proxies coevolve with their hosts, and insofar as the former increases the robustness of the latter’s capacities, proxies could result in system trait dependence and degeneration (\citep{edelmanDegeneracyComplexityBiological2001}).

SYSTEM EVOLUTION. In positing that intelligence equates with efficiency on
familiar goals and reasoning and invention on novel goals, TIS provides a
framework that can contribute to a greater understanding of system
assembly, growth, diversification and complexity (e.g., \citep{stanleyCompetitiveCoevolutionEvolutionary2004}; \citep{barrettModularityCognitionFraming2006}; \citep{triaDynamicsCorrelatedNovelties2014}; \citep{wongRolesFunctionSelection2023}; \citep{sharmaAssemblyTheoryExplains2023}). Empirical bases for change during lifetimes
and through generations include (1) on lifetime scales, intelligence is
influenced by sensing and past experience (information, knowledge,
skills) and influences current strategies and future goals (\citep{klyubinEmpowermentUniversalAgentCentric2005}; \citep{pearlBookWhyNew2018}), and (2) on multigenerational
evolutionary scales, natural selection drives intelligence trait
evolution in response to relative performance, needs and opportunities (\citep{pinkerCognitiveNicheCoevolution2010}; \citep{burkartEvolutionGeneralIntelligence2017}). There is also theoretical support that
systems evolve or coevolve in complexity with the relevance of
environmental challenges (\citep{adamiWhatComplexity2002}; \citep{klyubinEmpowermentUniversalAgentCentric2005}; \citep{frankNaturalSelectionMaximizes2009}; \citep{zamanCoevolutionDrivesEmergence2014}). In contrast, excepting for humans and select animals
(e.g., \citep{liTransformationsCouplingsIntellectual2004}; \citep{whitenEvolutionAnimalCultures2007}), we lack evidence for how
intelligence traits actually change during lifetimes given costs, benefits and tradeoffs, and suggest that
life-history theory could provide testable predictions (\citep{stearnsLifeHistoryEvolution2000}).

NICHE EVOLUTION. Depending on system type, goals in the intelligence niche might vary relatively little (e.g., prokaryotes) or considerably (e.g., mammals) in complexity. This means that a system with limited capacities will – in the extreme – either need to solve each of a myopic sequence of uncertainties, or use basic \textit{n}+1 planning abilities to achieve greater path manageability, efficiency, and goal accuracy and completeness. Under the reasonable assumption that systems can only persist if a sufficient number of essential (growth, survival, reproduction) goals are attainable, consistent with the ``competencies in development'' approach (\citep{flavellMetacognitionCognitiveMonitoring1979}), TIS predicts that simple cognitive solving precedes metacognitive planning, both during the initial development of the system and in situations where systems enter new intelligence niches. Moreover, TIS predicts that due to the challenges in accepting and employing proxies, the population distribution of proxy-supported intelligence should be initially positively skewed when a new proxy (e.g., Internet) is introduced, gradually shifting towards negative skew when the majority of the population has access to, needs and adopts the innovation (\citep{rogersDiffusionInnovations5th2014}).

Moreover, TIS places the notions of difficulty and surprisal in a new
perspective. To the extent that intelligence traits are typical of other
phenotypic traits, and even if an oversimplification, selection is expected to improve existing
intelligence capacities in familiar environments and diversify
intelligence capacities in novel environments. Both constitute the raw
material enabling successful encounters with new, more complex goals
(\citep{godfrey-smithEnvironmentalComplexityEvolution2002}). I predict that difficulty and surprisal continually
decrease in current intelligence niches, but increase as systems extend
existing niches or explore new niches.

TAXON EVOLUTION. The above predictions converge on the idea that solving comes before planning, but planning needs solving to be effective; that is, the latter cannot completely substitute for the former. This logic implies that solving evolution precedes the emergence of planning across phylogeny, that is, some degree of solving trait diversification would have occurred before the appearance of metacognitive (or more basal meta) abilities, and specifically planning. Nevertheless, if solving is ancestral and conserved as reactive functions or active behaviors, then transitions from B1 to B2 intelligence and from B2 to humans (see \hyperlink{tab1}{Table 1} legend for categories) would have implicated the emergence of augmented or novel solving abilities and planning capacities. For example, some cognitive adaptations in birds, mammals and cephalopods involve the development of brain regions permitting more acute perception, reaction to movement and anticipation of near future events (\citep{smithAnimalMetacognitionTale2014}; \citep{schnellHowIntelligentCephalopod2021}). Disentangling the selective sources (i.e., solving and/or planning) of generic ability evolution will be a daunting challenge. 
\\

\hypertarget{Ideas}{%
\subsection{Ideas}\label{Ideas}}

\emph{\textbf{Transitions in Intelligence?}}

Major taxonomic groups are differentiated by their capacities to resolve
their niche-relevant goals in niche-relevant environments (\citep{godfrey-smithEnvironmentalComplexityEvolution2002}). An interesting question that emerges from TIS is whether
the distribution of intelligence traits is continuous
across taxa, or rather shows distinct discontinuities with the emergence of new clades, suggestive of innovations or
transitions in intelligence. Although it is premature to claim the
distinctions in \hyperlink{tab1}{Table 1} have a solid scientific basis, I
speculate that evolutionary discontinuities in intelligence require
coevolution of intelligence systems with environments.
System-environment coevolution can be understood as follows.
Intelligence systems are characterized by abilities at different levels
(\hyperlink{tab1}{Table 1}) and the system infrastructure that supports these abilities
(e.g., in computing: operating systems; AI: large language models; biology: molecular signaling,
nervous systems). Some, possibly all, intelligence traits are labile to change and depending
on the trait and the system type, change can be due to individual experiences, associative learning, or molecular or cultural evolution. The extent to
which these and other sources of change occur will depend on contexts, i.e., interactions with
environments. Environments change and can ``evolve'' since they contain
the system population itself (intraspecific interactions), other species that
potentially evolve and vary in ecologies (interspecific interactions),
but also changing features
of the intelligence niche, including how the products of intelligence change the niche (e.g., construction and engineering; \citep{lalandIntroductionNicheConstruction2016}). Our understanding of system-environment
coevolution (\citep{brandonCoevolutionOrganismEnvironment1996}; \citep{adamiWhatComplexity2002}), system adaptation
to predictive cues, and evolution to predict environments (\citep{calvoConditionsMinimalIntelligence2015}; \citep{donaldson-matasciFitnessValueInformation2010}) is in its early days, but holds the potential for an
assessment of the richness of intelligence traits and how
intelligence ruptures and innovations may emerge.

One interesting question is whether discontinuities in intelligence are somehow associated with transitions in individuality. Conceptually,
collective intelligence is a form of proxy, suggesting
that the evolution of collective intelligence might in some instances
begin with individuals facultatively serving as intelligence proxies for others. Key
to collective intelligence is information sharing and goal alignment 
(\citep{kaufmannActiveInferenceModel2021}; \citep{leonardCollectiveIntelligencePublic2022}) and this supports the hypothesis that selection on intelligence traits when interests are aligned can result in transition to the higher individual, including intelligence traits that emerge at the collective level.

Moreover, should collective intelligence emerge and the collective be
sufficiently large and encounter sufficiently diverse challenges, then theory would suggest (\citep{ruefflerEvolutionFunctionalSpecialization2012}), division of labor in intelligence
should emerge. Division of labor can produce
faster and more efficient goal achievement and more diverse capabilities
permitting more goals. Thus, the outstanding
question that might lead to a theory of transitions in intelligence is
whether the evolution of higher levels of individuality are accompanied (or even driven) by innovations in intelligence. Nevertheless, to the extent that a macroevolutionary
theory of intelligence will be inspired by transitions in individuality (\citep{westMajorEvolutionaryTransitions2015}),
the theory would somehow need to incorporate division of labor in solving and planning (see \citep{gorelickNormalizedMutualEntropy2004} for information theory in social groups) and account for outliers that have not transitioned to stable collective units, notably certain mammals, bird taxa
and cephalopods (\citep{amodioGrowSmartYoung2019}).
\\

\emph{\textbf{From crystals to humans}}

CRYSTALS TO BIOLOGY. Uncertainty reduction and corresponding information
gain may involve one or more of ordering, mutating, morphing and
recombining code. In biological systems these transformations are usually associated with some
form of activity, for example searching for resources or avoiding
predators. 
In the absence of active, non-spontaneous behavior, intelligence minimally is information gain. Some of the simplest
reactions that result in reduced uncertainty and structured regularity
are the formation of inorganic crystals (e.g., \citep{frenkelOrderEntropy2015}). Purely
``physical intelligence'' in inorganic crystal formation employs
information (atoms or molecules) and processors (physical and chemical
charges and forces) in the context of surrounding environmental
conditions (pressure, temperature). To the extent that there is only a
single path from disordered mineral to ordered crystal, ``intelligence''
is the transition to atomic order. Given a
single deterministic reaction sequence, there can be no surprisal since there is no previous experience,
though some crystals may be more ``difficult'' to form than others due to
limiting environmental conditions and complex reaction sequences (e.g.,
\citep{gebauerCrystalNucleationGrowth2022}). These and other basic features of certain
physio-chemical systems have been hypothesized to have set the stage for
the emergence of biological life on Earth (\citep{morowitzEmergenceEverythingHow2002}), and may
comprise the scaffolding that permits certain cell
behaviors, such as membrane permeability, DNA expression and
replication, organelle function, protein synthesis and enzyme
behavior (\citep{smithOriginNatureLife2016}).

BIOLOGY. Taxonomic comparisons of intelligence across the tree of life
are largely meaningless in the absence of consideration of environments and species
niches. The types, diversities and amplitudes of intelligence traits as
habitat adaptations is an understudied area, with most work
focused on species with exceptional abilities from the standpoint of humans (\citep{rothConvergentEvolutionComplex2015}). TIS says that
intelligence capacities will be influenced by environmental conditions,
but gives no prediction regarding the relative fitness benefits of different traits. Rather, TIS says that difficulty is a feature of all of life, both because traits
are never perfectly optimized to all goal challenges and since trait
evolution always lags behind environmental change and changes in the niche
(\citep{smithOptimizationTheoryEvolution1978}). Intelligence
in B1 life involves the accurate sensing of the
environment and decision making (\citep{dussutourPhenotypicVariabilityPredicts2019}; \citep{calvoPlantsAreIntelligent2020}; \citep{s.reberSentientCellCellular2023}). Behavioral alternatives in, for example, metabolic
pathways and movement patterns, are observed across prokaryotes and
eukaryotes, although the difficulty involved is not in long-range future
planning, but rather current or short-range accurate and efficient solving (rate limitations
and tradeoffs among competing activities). In contrast, the notion of
surprisal in B1 taxa is more nuanced, since many species can exhibit
plastic responses to sufficiently infrequently encountered challenges
and opportunities (\citep{chevinEvolutionPhenotypicPlasticity2017}). An interesting possibility is
that plasticity is the most fundamental way to \emph{flexibly} deal with
uncertainty and novelty (\hyperlink{tab1}{Table 1}; \citep{godfrey-smithEnvironmentalComplexityEvolution2002}). That is, phenotypic and
behavioral plasticity in B1 taxa are at the foundation of intelligence in
these species and some of these traits are conserved across in B2 life and in humans.

BIOLOGY TO AI. Differences between biological intelligence and
artificial intelligence have been extensively discussed (e.g., \citep{sinzEngineeringLessArtificial2019}) the main distinctions including (1) the dependence of AI
functioning on non-AI sources (Memory and Humans as Proxies in \hyperlink{fig6}{Figure 6}) and
(2) limited capacities for inference in AI (Controller and Processor in \hyperlink{fig6}{Figure 6}).
Notably, since AI is a semi- or non-autonomous product of human ingenuity, AI functions as a proxy to human intelligence. TIS accounts for the
evolution of proxies (see also \citep{lachmannLifeAlive2019a}), that is new functions in technology that augment
or replace existing human capacities. As AI becomes
increasingly independent and achieves AGI, humans themselves might
function as proxies for the technology they created.

VOLUTION. Intelligence functions not only to
achieve goals, but as experiences that promote future improvements to
accuracy and efficiency and the latitude to engage in new goal spaces.
Importantly, experiences are not always successful and learning from
errors is well-recognized as improving performance
(\citep{shadmehrErrorCorrectionSensory2010}; \citep{metcalfeLearningErrors2017}). As such, I speculate that in addition to successful outcomes,
inefficient, erroneous or dead-end paths contribute information that can
be recycled or repurposed to undertakings not necessarily
affecting Darwinian fitness, such as leisure, play, politics, games and
art (see also \citep{chuPlayCuriosityCognition2020}). If correct, then TIS is applicable to
both theories of evolution (genetic, cultural, technological) and forms of
goal attainment not directly under natural selection.

HUMANS. A tremendous amount of research has been dedicated to the
``uniqueness'' of \emph{Homo sapiens} with numerous hypotheses and
theories on how and why humans have certain highly developed capacities
that are either less developed or non-existent in other species. Empirical
evidence is clear that modern humans have achieved unparalleled
(1) cognitive and communication abilities, and (2) cumulative culture and social,
technological and informational developments (e.g., \citep{gibsonToolsLanguageIntelligence1991}; \citep{tennieRatchetingRatchetEvolution2009}; \citep{pagelAdaptedCulture2012}; \citep{pinkerCognitiveNicheCoevolution2010}). Despite longevity (\citep{kaplanEmergenceHumansCoevolution2002}) and environmental challenges (\citep{kaplanTheoryHumanLife2000}) being
hypothesized as key components in the evolution of human intelligence (and see similar arguments for other animals, \citep{amodioGrowSmartYoung2019}),
cognitive and communication (language) capacities associated with
specific regions of the human brain and vocal apparatus are the \textit{only unique body systems} explaining stand-alone differences
between humans and other animals (\citep{lalandUnderstandingHumanCognitive2021}; \citep{gibsonEvolutionHumanIntelligence2002}).
Cumulative culture, on the other hand, could be abstractly regarded as a
group proxy, that is the storage, communication and expression of
cultural artefacts, including priors and knowledge, and higher-level embodiments such as social norms and
institutions. Culture and brains are hypothesized to have coevolved (\citep{muthukrishnaCulturalBrainHypothesis2018}), which suggests that human intelligence owes
to both the slow molecular evolution (brains) and the faster evolution
of information (cultures, individual information). From an individual
human standpoint, culture is a gifted information proxy from the group
that can increase intelligence (\citep{sternbergCultureIntelligence2004}), and indeed
this could be a great part of what distinguishes human from other-primate
intelligence (\citep{tomaselloCulturalTransmissionView2001}). 

HUMANS AND PROXIES. As argued above, I believe that proxies are central to the scaffolding, amplification and diversification of
human intelligence. This is consistent with ideas that proxies and
culture coevolve (\citep{brinkmannMachineCulture2023}), and that the limit to the individual can
extend to proxies (\citep{krakauerInformationTheoryIndividuality2020}) and the emergence of a new level
of individuality (\citep{raineyMajorEvolutionaryTransitions2023}). TIS has the ingredients to explain the
generality of human intelligence, but would need to be further developed into
explicit mathematical models in order to understand the contributions of
individual systems and proxies such as culture. Individual brains, group
culture and technology evolve on different time scales (\citep{perreaultPaceCulturalEvolution2012}) and the fact that innovations such as AI are developing so rapidly
is consistent with the idea that evolutionary runaways are possible (\citep{leeIdeaEnginesUnifying2024}). Thus, a tantalizing possibly to explain the
discontinuity of human intelligence with respect to other biological
systems is that coevolution of systems and their constructions in the intelligence niche
is an auto-catalytic process (e.g., \citep{cazzollagattiEmergenceEcologicalEconomic2020}), producing the intelligence runaway that is arguably occurring in human-proxy systems.
\\

\hypertarget{Acknowledgements}{%
\subsection{Acknowledgements}\label{Acknowledgements}}
I thank Steve Frank, David Krakauer, Michael Lachmann, Julien Luneau, and Melanie Mitchell for discussions.

\newpage

\hypertarget{tab1}{Table 1}. TIS is based on functions and associated abilities. Operators, instruments and optimizers are divided into ability levels, hypothetically associated with evolutionary complexity. Each ability could have variants of at multiple levels, and the table only associates each with their lowest hypothetical level. Abilities included in TIS equations are indicated in italics. System types associated with each feature are given, with uncertain or partial assignments indicated by parentheses. P=physical; AI=artificial intelligence; B1=unicellular, multicellular fungi and plants, invertebrate animals with the exceptions of cephalopods, some social microbes, and some social insects; B2=vertebrates and aforementioned exceptions of B1; H=humans. See main text for further details and explanations.
\begin{figure}		
    \begin{center}	
    \includegraphics[width=1.0\textwidth]{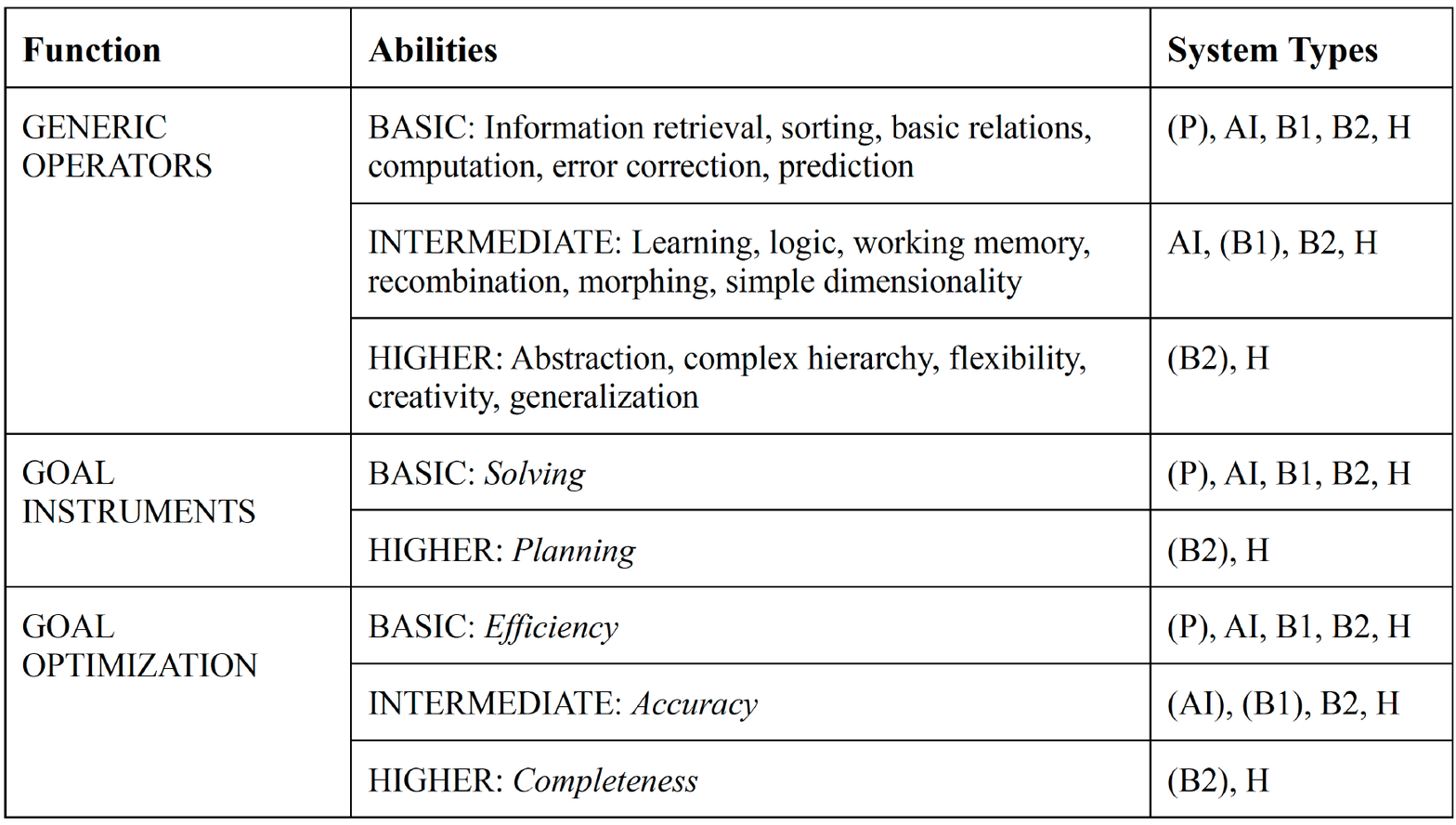}
    \end{center}
\end{figure}\normalsize

\newpage

\begin{figure}		
    \begin{center}	
    \includegraphics[width=1.0\textwidth]{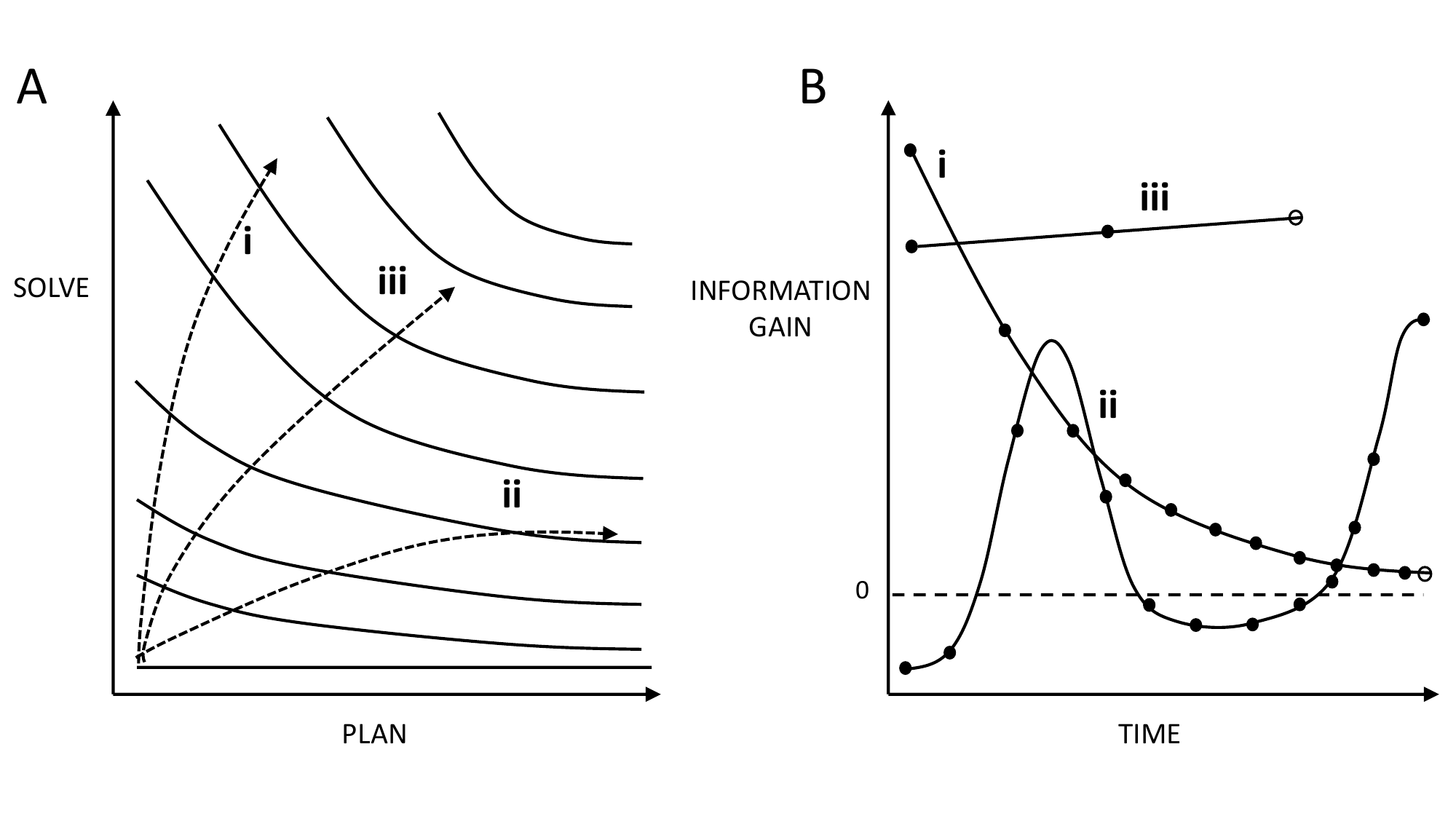}\\
    \end{center}
\end{figure}\normalsize
\hypertarget{fig1}{Figure 1}. A -- Key concepts in the Theory of Intelligences. Solving is the ability to resolve uncertainty. Planning is the ability to trace a future sequence of manageable challenges towards goal resolution. The clines refer to progressive intelligence milestones in abilities to deal with complexity. Trajectory i is gains in intelligence mostly due to solving, whereas ii is dominated by planning. Trajectory iii, which initially favors solving and gradually favors planning is the most efficient of the three examples, leading to higher marginal and overall gains in milestones. Note that the three trajectories are different rotations of the same curve. B -- Time courses of path node changes for three hypothetical scenarios. Each point (node, subgoal) corresponds to a goal-related information gained in the path (represented by a continuous line). Goal resolution is indicated by an open circle. i: Monotonic decrease in path node changes leading to an intermediate resolution time (efficiency). ii: Complex trajectory in path node change, with transient information loss, indicative of the discovery of a more accurate path and a consequential delay in (off graph) resolution. iii: Efficient path corresponding to a rapid resolution.

\newpage

\begin{figure}		
    \begin{center}	
    \includegraphics[width=1.0\textwidth]{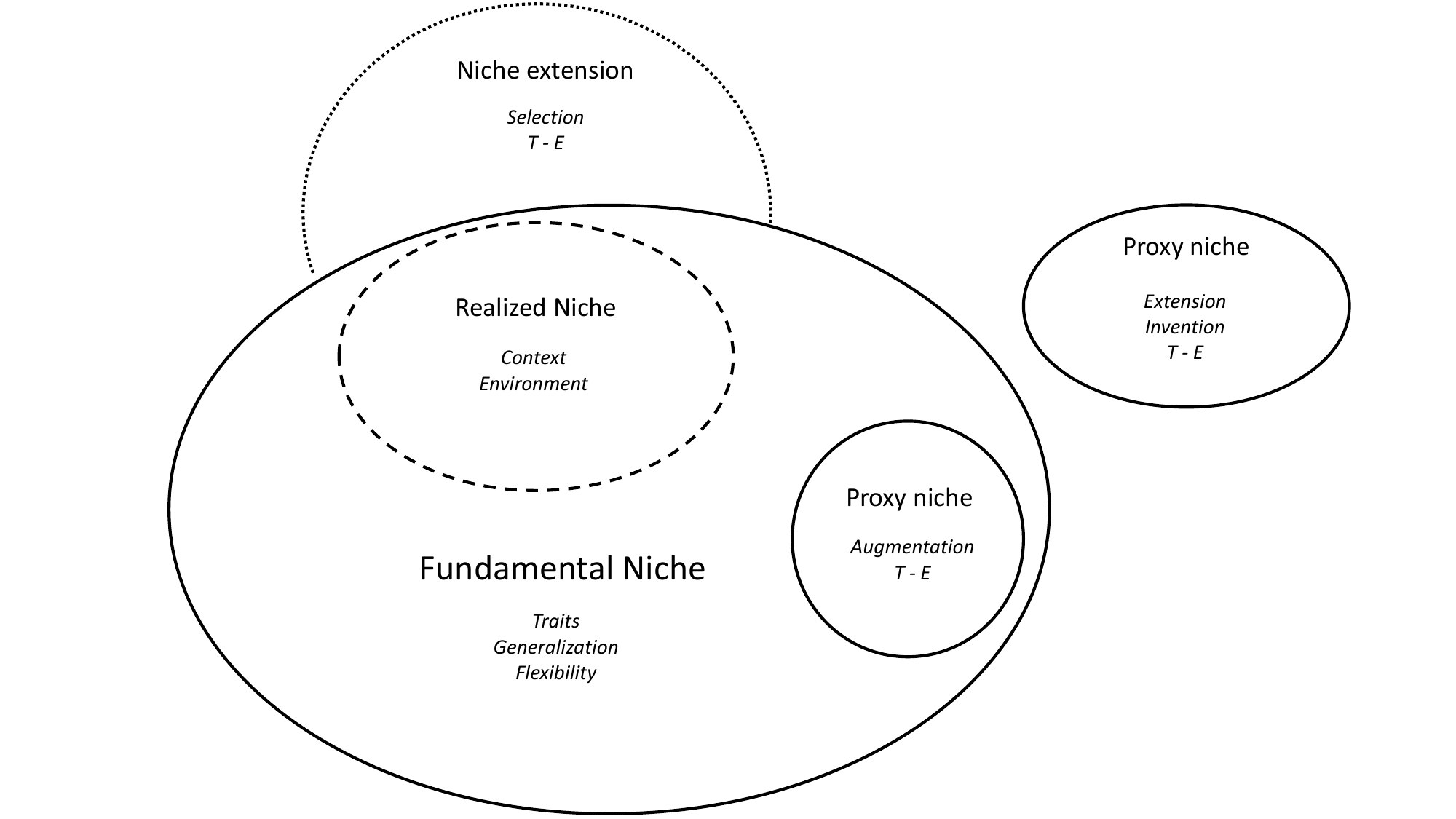}\\
    \end{center}
\end{figure}\normalsize
\hypertarget{fig2}{Figure 2}. The intelligence niche. Shown are representations borrowed from the ecological and evolutionary sciences, whereby the \textit{fundamental niche} of a system (or a species) is the set of possible conditions for which the population can persist. Conditions include contexts, environments and goal types. The \textit{realized niche} is the set of conditions under which the individual or population actually operates due to limited contexts, goal types/choices, and environments experienced. The fundamental intelligence niche can be extended via some combination of transmission (learning, experience) or system evolution. Proxies can extend the realized niche, either within the fundamental niche (augmentation of existing capacities) or through the invention of new niches. New niches therefore emerge as a system innovation (e.g., cognition, language...) or extension of the individual system (e.g., social, technology...), possibly resulting in a discontinuity or transition in intelligence. T - E: transmission and evolution.

\newpage

\begin{figure}		
    \begin{center}	
    \includegraphics[width=1.0\textwidth]{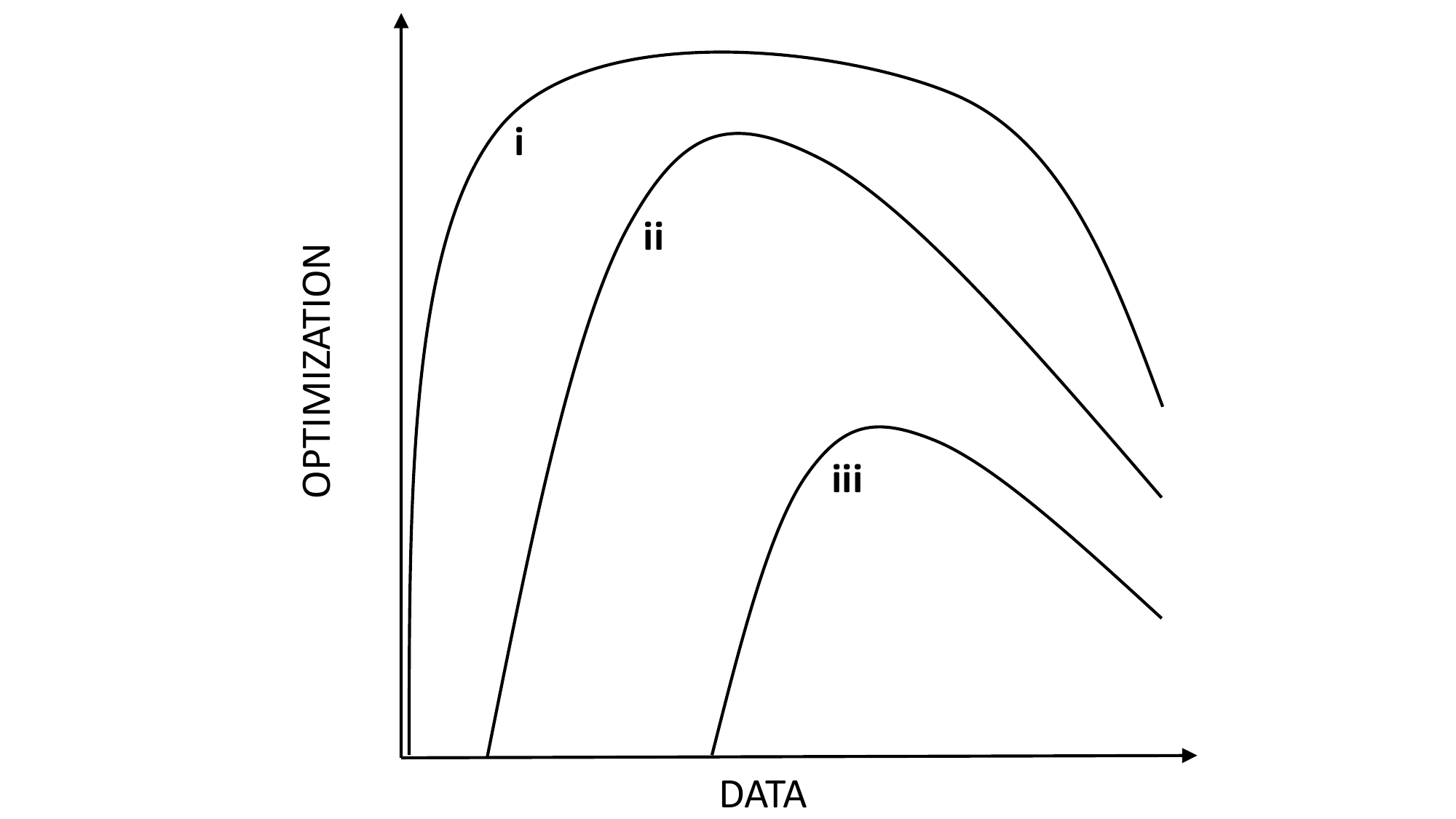}\\
    \end{center}
\end{figure}\normalsize
\hypertarget{fig3}{Figure 3}. The hypothetical effects of data richness on the optimization (efficiency, accuracy, completeness) of a resolution to a goal. Insufficient data lowers optimality. Too much data compromises abilities to optimally parse information. Curves i – iii represent goals of increasing complexity and difficulty. Simple goals can be optimized to higher levels with less data than more complex goals. Curves i – iii can also be interpreted as decreasing intelligence profiles across individual systems.

\newpage

\begin{figure}		
    \begin{center}	
    \includegraphics[width=1.0\textwidth]{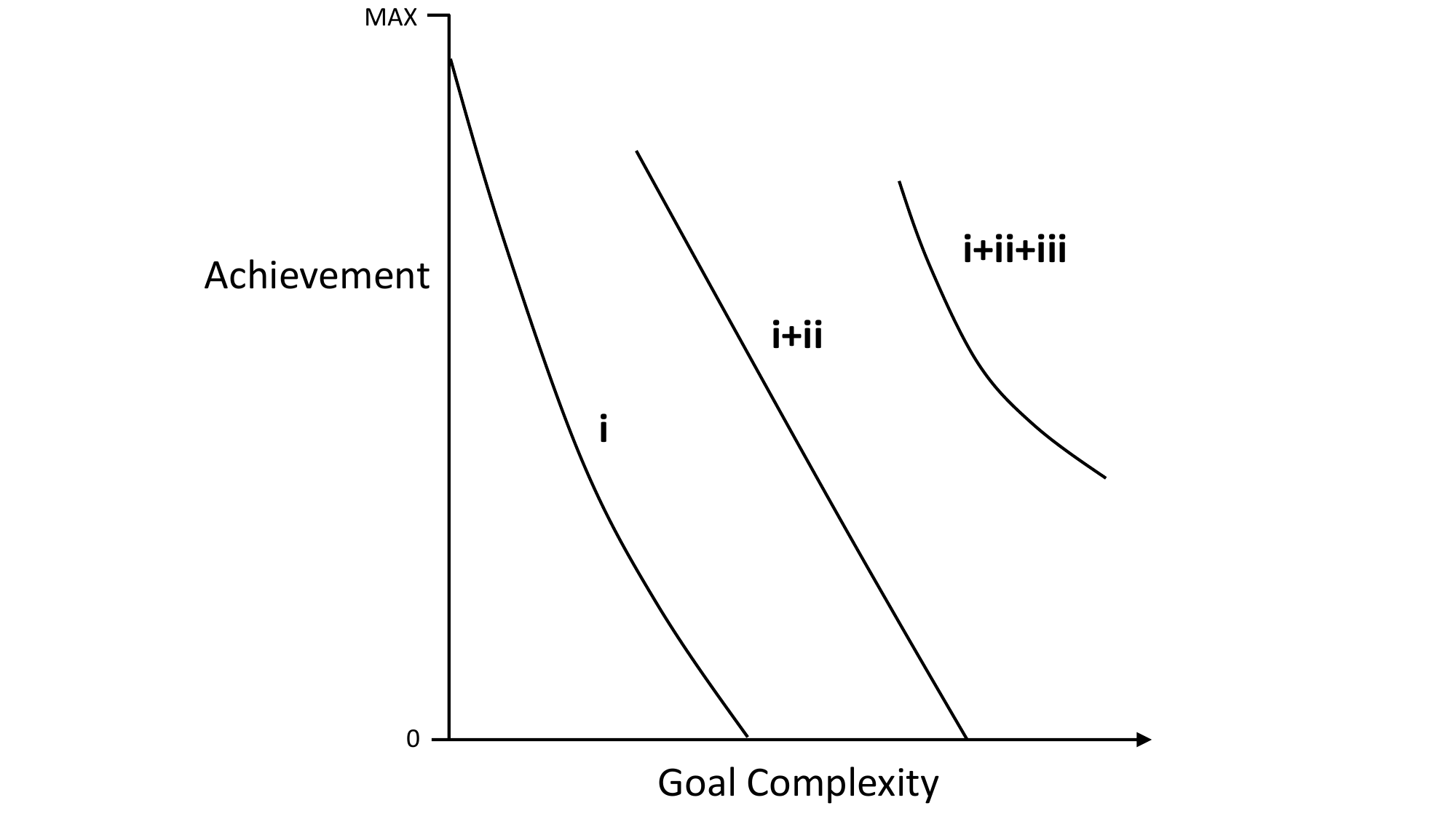}\\
    \end{center}
\end{figure}\normalsize
\hypertarget{fig4}{Figure 4}. Hypothetical representation of how complexity in intelligence traits respond to complexity in goals. Curve i corresponds to solving abilities only, which are effective over a limited range of lower goal complexities. Curve i+ii corresponds to the emergence of basic-level planning together with existing solving, which increases capacities compared to solving alone. Curve i+ii+ii is higher level planning at sufficiently high goal complexity, together with basic-level planning and solving. Achievement is some combination accuracy, completeness and efficiency. Assumed is that higher levels only emerge when scaffolding already exists and evolutionary benefits of innovations exceed costs. See main text for discussion. 

\newpage

\begin{figure}		
    \begin{center}	
    \includegraphics[width=1.0\textwidth]{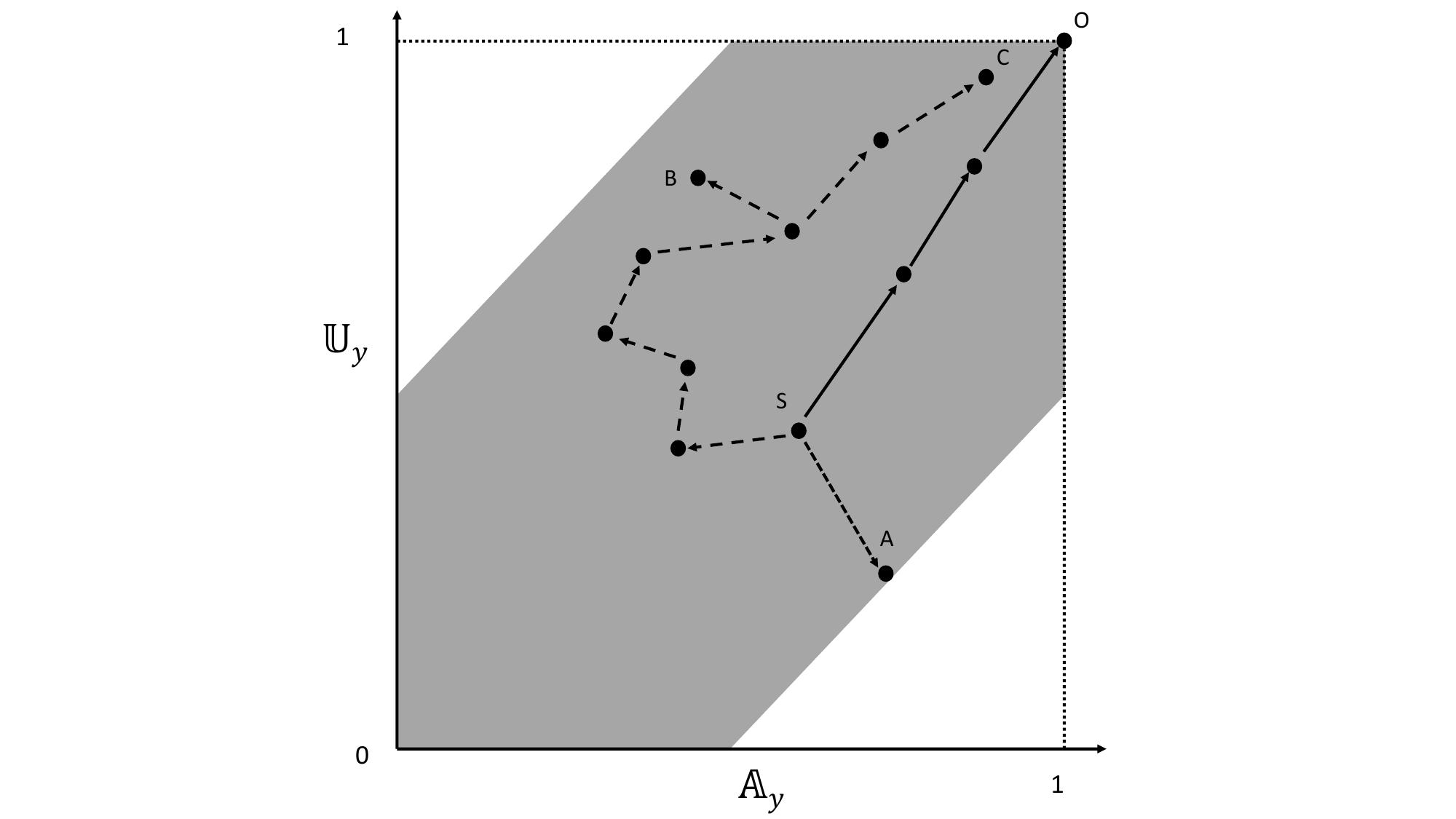}\\
    \end{center}
\end{figure}\normalsize
\hypertarget{fig5}{Figure 5}. Hypothetical examples of sequential changes in solving \(\mathbb{U}_{y}\) and planning \(\mathbb{A}_{y}\) in goal resolution, based on equations (1b) and (2b). The maximally accurate and complete solution is denoted O. Path trajectories vary in terms of abilities to solve and plan, as reflected by the direction and length of arrows and the number of nodes from start (S) to resolution (A, B, C or O). The segments from S to O reflect an efficient, accurate and complete resolution. The single jump trajectory to a poor resolution at A (lower information, slightly greater planning accuracy) is due to low solving ability. The comparatively greater number of smaller segments from S to the sub-optimal resolution at B is due in particular to low planning ability. These examples are oversimplifications and intended to illustrate basic principles of TIS. See main text for further explanation.

\newpage

\begin{figure}		
    \begin{center}	
    \includegraphics[width=1.0\textwidth]{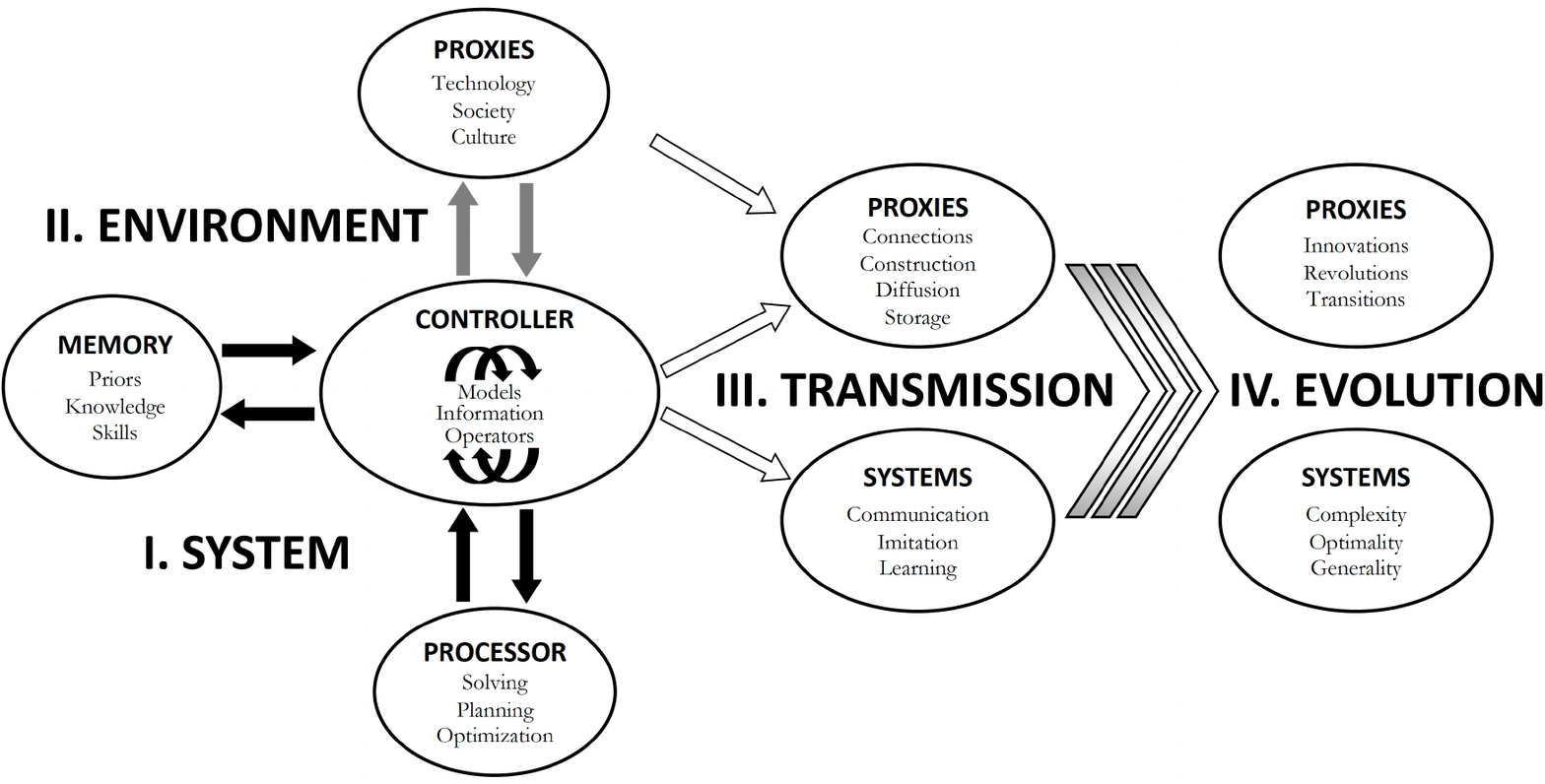}\\
    \end{center}
\end{figure}\normalsize
\hypertarget{fig6}{Figure 6}. Schematic diagram of TIS in the intelligence ecosystem. The SYSTEM phenotype (I) is composed of the CONTROLLER, PROCESSOR and MEMORY. The CONTROLLER generates the goal-directed information based on generic abilities (Generic Operators, Table 1), and how these abilities are instrumentalized via the PROCESSOR, which solves uncertainty and plans and optimizes paths to goal resolution. The SYSTEM interacts with (II) the ENVIRONMENT both in setting and addressing GOALS. The extended phenotypes are PROXIES such as social interactions, technology and culture. Intelligence can be codified \textit{in} the SYSTEM as phenotypic traits, stored MEMORY, and hard-wired or plastic behaviors. Intelligence can also be codified \textit{outside} the SYSTEM in (non-mutually exclusive) PROXIES, such as collectives, society, culture, artefacts, technology and institutions. The current SYSTEM (I and II) is based on past and current TRANSMISSION and intelligence trait EVOLUTION (not shown), and develops and integrates intelligence traits over the SYSTEM’s lifetime (not shown), and influences future (III) TRANSMISSION and (IV) EVOLUTION. The intelligence niche is affected by SYSTEM EVOLUTION and possibly SYSTEM-dependent PROXY EVOLUTION and this occasionally produces intelligence innovations and more rarely, intelligence transitions. 

\newpage

\bibliography{Intelligence2024}

\end{document}